\documentclass[conference]{IEEEtran}

\usepackage{amssymb}
\usepackage{graphicx}
\usepackage{amsmath}
\usepackage{algorithm}
\usepackage{algorithmic}
\usepackage{mathtools}
\usepackage{stmaryrd}
\usepackage{subfigure}
\usepackage{adjustbox}
\usepackage{multirow}
\usepackage{enumitem}
\usepackage{booktabs} 
\usepackage{url}
\usepackage{tcolorbox}
\usepackage{amsthm}
\usepackage{balance}
\usepackage{flushend}

\newtheorem{definition}{Definition}

\begin{document}
\title{Data-driven Blockbuster Planning on Online Movie Knowledge Library}

\let\proof\relax
\let\endproof\relax

\setlength{\abovetopsep}{0.5ex}
\setlength{\belowrulesep}{0pt}
\setlength{\aboverulesep}{0pt}

\newcommand{\sur}[1]{\ensuremath{^{\textrm{#1}}}}
\newcommand{\sous}[1]{\ensuremath{_{\textrm{#1}}}}

\newcommand{\ie}[0]{\textit{i.e.}}
\newcommand{\eg}[0]{\textit{e.g.}}
\newcommand{\etc}[0]{\textit{etc.}}
\newcommand{\problem}[0]{$\text{BP}$}
\newcommand{\ours}[0]{$\text{BigMovie}$}
\newcommand{\cast}[0]{production team}
\newcommand{\castb}[0]{Production Team}


\widowpenalty = 10000

\author{
    \IEEEauthorblockN{Ye Liu\IEEEauthorrefmark{1}, Jiawei Zhang\IEEEauthorrefmark{2}, Chenwei Zhang\IEEEauthorrefmark{1}, Philip S. Yu\IEEEauthorrefmark{1}\IEEEauthorrefmark{3}}
    \IEEEauthorblockA{\IEEEauthorrefmark{1}Department of Computer Science, University of Illinois at Chicago, IL, USA
    \\\{yliu279, czhang99, psyu\}@uic.edu}
     \IEEEauthorblockA{\IEEEauthorrefmark{2}Department of Computer Science, Florida State University, FL, USA
    \\\{jzhang\}@cs.fsu.edu}
    \IEEEauthorblockA{\IEEEauthorrefmark{3}Institute for Data Science, Tsinghua University, Beijing, China}
}

\maketitle

\begin{abstract}
In the era of big data, logistic planning can be made data-driven to take advantage of accumulated knowledge in the past. While in the movie industry, movie planning can also exploit the existing online movie knowledge library to achieve better results. 
However, it is ineffective to solely rely on conventional heuristics for movie planning, due to a large number of existing movies and various real-world factors that contribute to the success of each movie, such as the movie genre, available budget, production team (involving actor, actress, director, and writer), etc. In this paper, we study a ``Blockbuster Planning'' ({\problem}) problem to learn from previous movies and plan for low budget yet high return new movies in a totally data-driven fashion. After a thorough investigation of an online movie knowledge library, a novel movie planning framework ``\underline{B}lockbuster Plann\underline{i}n\underline{g} with \underline{M}aximized M\underline{ov}ie Conf\underline{i}guration Acquaintanc\underline{e}'' ({\ours}) is introduced in this paper. From the investment perspective, {\ours} maximizes the estimated gross of the planned movies with a given budget. It is able to accurately estimate the movie gross with a $0.26$ mean absolute percentage error (and $0.16$ for budget). Meanwhile, from the production team's perspective, {\ours} is able to formulate an optimized team with people/movie genres that team members are acquainted with. Historical collaboration records are utilized to estimate acquaintance scores of movie configuration factors via an acquaintance tensor. We formulate the {\problem} problem as a non-linear binary programming problem and prove its NP-hardness. To solve it in polynomial time, {\ours} relaxes the hard binary constraints and addresses the {\problem} problem as a cubic programming problem. 
Extensive experiments conducted on IMDB movie database demonstrate the capability of {\ours} for an effective data-driven blockbuster planning.
\end{abstract}

\IEEEpeerreviewmaketitle

\section{Introduction}\label{sec:introduction}
The movie industry attracts great interests from both movie investors and the public audience because of its high profits and entertainment nature. Attracted by the huge market, lots of investors are inquiring about identifying high-gross movies to invest in. Besides recognizing profitable movies, it is rewarding to provide a reasonable and promising planning for a new movie at its developmental stage, which has been greatly ignored in previous works due to the complexity of various factors, including the movie genre and {\cast} (actor, actress, writer and director).


The booming movie industry has accumulated thousands of previous movies as well as their gross statistics, which may serve as a movie knowledge library to help achieve better results for future movie planning. Therefore it is no longer efficient to rely on conventional heuristics for comprehensive movie planning \cite{fowler2008heuristics}. Data-driven movie planning methods are in great need to exploit the accumulated knowledge to support the decision-making process when planning for a new movie. The data-driven planning has shown a huge success on the well-known TV series ``House of Cards'', produced by Netflix, using the data collected from viewer\footnote{\url{https://thenextweb.com/insider/2016/03/20/data-inspires-creativity/}}.

Generally, popular movie genres and renowned movie stars are the favorable choices during the planning so as to maximize the gross. But remuneration of the movie stars and movie's available budget also need to be considered in the movie planning. Meanwhile, a seamless collaboration among team members is the premise of high gross. For example, it will always be easier for directors to continue working on a new movie with the movie genre and production team members that they are acquainted with. And the old acquaintances can always have a tacit understanding and easy to arouse spark when they cooperate in their new movies. 


\textbf{Problem Studied:} In this paper, a research problem, namely the ``\underline{B}lockbuster \underline{P}lanning'' ({\problem}) problem, is introduced. Given an online movie knowledge library which consists of the existing movie information, we plan the movie configuration including genre and production team for a new movie under a pre-specified budget. We note that although there are occasions where a low budget production with unknown stars becomes a hit, we focus on the common cases involving known persons with available data. The objective of an optimal planning is to achieve: (1) the maximized gross, and (2) the optimized acquaintance among the movie configuration factors. 


The {\problem} problem studied in this paper is a novel research problem, and few existing methods can be applied to solve it. The {\problem} problem significantly differs from related works, such as $(1)$ \textit{movie gross prediction} \cite{mestyan2013early}, $(2)$ \textit{viral marketing} \cite{richardson2002mining,kempe2003maximizing}, $(3)$ \textit{team formation} \cite{lappas2009finding,anagnostopoulos2012online}. $(1)$ The \textit{movie gross prediction} problem \cite{sharda2006predicting} studied in existing works merely focuses on inferring the movie gross while the {\problem} problem aims at providing the optimal planning of various movie factors which can lead to the optimal gross for investors. $(2)$ {\problem} and the \textit{viral marketing} problems \cite{kempe2003maximizing} are both planning problems that aimed at maximizing certain target objectives, but they are solving totally different problems in distinct scenarios: $a)$ \textit{viral marketing} problems are usually studied in online social networks based on certain information diffusion models, while the {\problem} problem is studied in the online movie knowledge libraries instead; $b)$ \textit{viral marketing} problems aim at maximizing the infected users, while {\problem}'s objective is to maximize the movie gross; $c)$ instead of selecting the optimal users in \textit{viral marketing} problems, the {\problem} problem aims at planning for an optimal movie factor configurations. Recently, a variation of the LT model named PNP \cite{Koutra2017} is proposed for the movie design problem. The objective of PNP is very similar to our work except that PNP aims to attract most of the target users but our model aims to achieve the maximum gross under the given budget. $(3)$ Different from conventional \textit{team formation} problems \cite{lappas2009finding}, where team members are planned for the entrepreneurial team project base on satisfying skill qualification and minimizing the communication cost of the team members, our method also aims to maximize movie gross. 

The BP problem is challenging to solve due to:
\begin{itemize}
\item \textit{Unknown Movie Success Factors}: What are the contributing factors in the success of a movie? Few research works have ever been studied this problem, and relevant movie factors are still unknown. 

\item \textit{Movie Gross/Budget Function}: How much gross (budget) can a movie make (require), given a configuration of the movie success factors? A proper estimation of the movie gross and budget will be required for studying the {\problem} problem.

\item \textit{Movie Configuration Acquaintance Function}: How to compute the acquaintance scores among the movie configuration factors? A function that can measure acquaintance properly is needed in defining the {\problem} problem.

\item \textit{NP Hardness}: Based on our analysis, we demonstrate that the BP problem is actually an NP-hard problem, and no solution exists that can solve it in polynomial time if P $\neq$ NP. 

\end{itemize} 
To solve the aforementioned challenges, a new movie planning framework ``\underline{B}lockbuster Plann\underline{i}n\underline{g} with \underline{M}aximized M\underline{ov}ie Conf\underline{i}guration Acquaintanc\underline{e}'' ({\ours}) is proposed in this paper. With a thorough analysis of an online movie knowledge library dataset, IMDB, a set of factors affecting movie success are identified. The effectiveness of these extracted factors are validated in Section \ref{sec_ver}. The acquaintance scores of the movie configuration factors can be calculated based on an acquaintance tensor constructed with the historical collaboration records which is discussed in great detail in Section \ref{sec_plan}. The {\problem} problem is formulated as a constrained optimization problem with hard binary constraints, which aims at maximizing the inferred gross function as well as the acquaintance measure. We further demonstrate that {\problem} is at least as difficult as the \textit{Knapsack problem} and the \textit{Maximal Clique problem}, which renders the BP problem to be NP-hard as well. By relaxing the hard constraints, we introduce an approximation solution to resolve the problem in polynomial time. For the experimental result, we can see BigMovie outperforms the competitors. In addition, at the end of the paper, the case study is provided, which demonstrate that by using {\ours}, a lucrative movie planning can be achieved.

\section{PROBLEM FORMULATION} \label{sec_formulation}
In this section, we will first define several important concepts used in this paper, and then provide the formulation of the {\problem} problem.
\subsection{Notation}
At the beginning of this section, we will first define some notations used in this paper. Throughout this paper, we will use lower case letters (e.g., $x$) to denote scalars, lower case bold letters (e.g., \textbf{x}) to denote column vectors, upper case letters (e.g., $X$) to denote elements of matrices, upper case calligraphic letters (e.g., $\mathcal{X}$ ) to denote sets, and bold-face upper case letters (e.g., \textbf{X}) to denote matrix and high-order tensors. $T$ is used to represent the transpose of a vector (e.g., $\bf{x}^T$). $||\cdot||_1$ is the $\ell_{1}$-norm of vector (e.g., $||\bf{x}||_1$).

\subsection{Terminology Definition}
\begin{definition} \textbf{Online Movie Knowledge Library}: An \emph{online movie knowledge library} can be represented as an undirected graph $G=(\mathcal{M}$,~$\mathcal{C}$,~$\mathcal{E}$,~$\mathcal{A})$, where node set $\mathcal{M}=\{m_1,~m_2,~...,~m_n\}$ denotes the set of n movies in the library and $\mathcal{C}=\{c_1,~c_2,~...,~c_l\}$ is the set of $l$ production team members. The node set $\mathcal{C}$ can be divided into $\mathcal{C}^t \cup~\mathcal{C}^s \cup~\mathcal{C}^w \cup~\mathcal{C}^d$, which denote the set of actors, actresses, writers and directors, respectively. Link $\mathcal{E}$ represents the relationship between movie production team and movies. For instance, link ($(c_i,m_j) \in \mathcal{E}$) indicates participation of a production team member $c_i$ in a movie $m_j$. And set $\mathcal{A}$ denotes the attribute of node set $\mathcal{M}$. For the movie $m_i$, the relative attribute is $\mathcal{A}_{(m_{i})}= \mathcal{A}_{(m_{i})}^g \cup \{a_{(m_{i})}^b, a_{(m_{i})}^g\}$  , where $\mathcal{A}_{(m_{i})}^g$ is the genre list of movie $m_i$, $a_{(m_{i})}^b$ is the budget of movie $m_i$ and $a_{(m_{i})}^g$ is the gross of movie $m_i$.
\end{definition}

\begin{definition} \textbf{Movie Configuration}: Each movie $m_i\in\mathcal{M}$ in the online knowledge library will have an unique configuration , covering \textit{movie production team} (involving \textit{actor}, \textit{actress}, \textit{writer} and \textit{director}), \textit{movie genre}, etc, which can be represented as vector $\textbf{x}_{(m_{i})}$=$[\textbf{x}_{(m_{i})}^t,~\textbf{x}_{(m_{i})}^s,~\textbf{x}_{(m_{i})}^d,~\textbf{x}_{(m_{i})}^w$ $, \textbf{x}_{(m_{i})}^g]$ $\in \mathbb R^{1\times N}$, where $\textbf{x}_{(m_{i})}^t$ represents the list of all actor, $\textbf{x}_{(m_{i})}^s$ is the list of all actress, $\textbf{x}_{(m_{i})}^w$ represents the list of all writer, $\textbf{x}_{(m_{i})}^d$ represents the list of all director and $\textbf{x}_{(m_{i})}^g$ represents the list of all genre of a movie $m_i$. $N$ is the sum of length of those lists. We will provide detailed representations in Section 4.1.
Besides those factors covered in the above \textit{movie configuration} definition, various other relevant factors (e.g., \textit{movie language}, \textit{production country}, etc.) can also be effectively incorporated with a simple extension to the definition, which will not be studied in this paper.
\end{definition}

\begin{figure*}[htbp!]
\centering
\includegraphics[width=18cm]{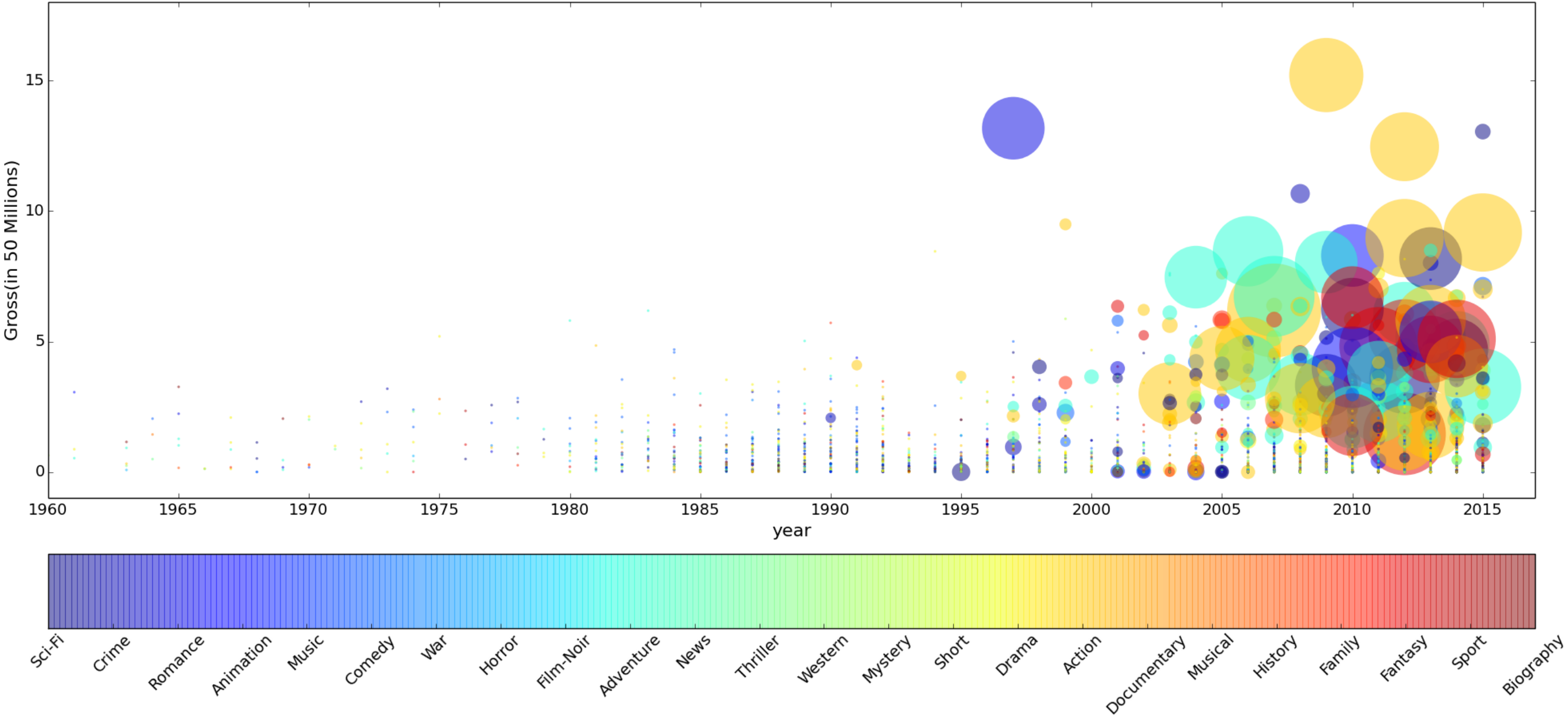}
\caption{Movie General information of IMDB Movie}
\label{fig:genral}
\end{figure*}

\begin{definition} \textbf{Movie Configuration Acquaintance}: Given two movie team members $c_i, c_j \in \mathcal{C}, c_i \neq c_j$ and a movie genre $g_k \in \mathcal{A}_{(m_{i})}^g$, their acquaintance can be represented as $Acquaintance(c_i,~c_j,~g_k)$, denoting their historical collaboration frequency. For instance, if crews $c_i$ and $c_j$ participate $t$ times in $g_k$ genre movie, $Acquaintance($ $c_i,~c_j,~g_k)=t$.
\end{definition}

\subsection{Problem Formulation}
\begin{definition} \textbf{Blockbuster Planning Problem}:
Given a fixed budget $B$, the objective of {\problem} is to plan a movie configuration $\textbf{x}$ that achieves maximum movie gross and maximum movie configuration acquaintance simultaneously, subject to the budget $B$. 

Let $Budget(\textbf{x})$ denotes the cost by using movie configuration $\textbf{x}$, $Gross(\textbf{x})$ estimates the gross earned by using $\textbf{x}$ and $Acquaintance(\textbf{x})$ measures the acquaintance of movie configuration. Formally, the BP problem aims at inferring the optimal movie configuration $\textbf{x}^*$ which can maximize the following objective function
\begin{align}
\textbf{x}^*=&\arg \max_{\textbf{x}}\alpha \cdot Gross(\textbf{x})+\beta \cdot Acquaintance(\textbf{x}) ,\\
s.t.& ~ Budget(\textbf{x}) \le B  \nonumber
\label{eq:obj}
\end{align}
\end{definition}
In above equation, the concrete representation of function $Budget(\textbf{x})$ and $Gross(\textbf{x})$ will be provided in Section \ref{sec_ver} and function $Acquaintance(\textbf{x})$ will be provided in Section \ref{sec:pref}. $\alpha$ is the coefficient of $Gross(\textbf{x})$ and $\beta$ is the coefficient of function $Acquaintance(\textbf{x})$. Analysis of parameters $\alpha$ and $\beta$ will be provided in the Section \ref{sec_exp}.

\section{ONLINE MOVIE KNOWLEDGE LIBRARY STATISTICAL ANALYSIS}\label{sec_anly}
 Before introducing the method to solve the blockbuster planning problem, in this section, we first study the IMDB \footnote{\url{http://www.imdb.com}} datasets to provide some statistical analysis about the factors affecting movie gross. The analysis of the IMDB movies focuses on several important aspects like the gross, budget, genres and production team information (Actor, Actress, Director, Writer), which provides fundamental insights for the blockbuster planning framework. Among the crawled IMDB movies, only $3,156$ movies contain the gross and budget information, and they belong to $24$ genres and cover $72,786$ actors, $38,951$ actresses, $4,576$ writers and $1,682$ directors.

\begin{figure*}[htbp!]
\centering
\includegraphics[width=18cm]{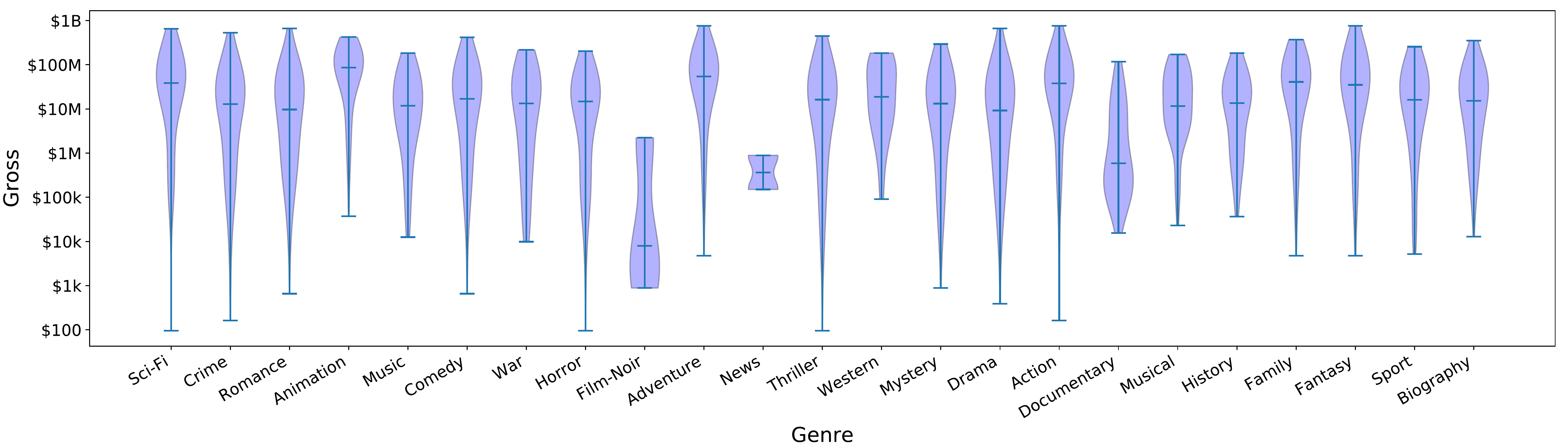}
\caption{Movie Gross Distribution of IMDB Movie Genres}
\label{fig:genre}
\end{figure*}

\subsection{General Movie Information Statistics}
In this section, we study general information, like budget and genre regarding the movie gross.

The results are shown in Figure \ref{fig:genral}. In this figure, we provide the information distribution of IMDB datasets in terms of their production years. In the plot, each circle denotes a movie, whose x axis and y axis denote the movie gross and the production year, respectively. Meanwhile, the circle diameter represents the budget of the movies (larger circle corresponding to movies with bigger budgets). Additionally, we use different colors, shown in the color bar below the figure, to represent the corresponding genre of each movie. 

According to Figure \ref{fig:genral}, we observe that the number of movies produced in recent years are increasing. For instance, according to our dataset, the number of movies produced in years 1980, 1990, 2000, 2010, and 2015 are 12, 58, 77, 137 and 158 respectively. Besides the movie numbers, we also discovered several important observations regarding the movie budget, gross and genre information.

\textbf{Recent Movies Have Higher Budget And Gross:} 
According to the movie budget data, a majority of the movies produced before 2000 have budgets under $\$$200 million while recent movies have relatively higher budgets (i.e., the circles in recent years are much larger). Among the top ten movies receiving the highest budget, six of them were produced within the past five years. Simultaneously, the movie gross of the past ten years is much higher than before (i.e., dot in recent years are much higher). Few of movies produced before 2000 had a gross of more than $\$$250 million while some movies produced after 2000 reached more than $\$$750 million on gross, which shows the growth of the movie industry. Among top ten highest gross movies, five of them were produced within the past five years, for example movie ``Jurassic World'' ($\$$652 million), ``The Avengers'' ($\$$623 million) and ``The Hunger Games: Catching Fire'' ($\$$424 million). The movie ``Avatar'' (produced in 2009) achieves the highest gross in our dataset, which is $\$$760 million. Additionally, the growth of the movie industry brings a big gross discrepancy from the recent movies, because the gross variance between movies produced before 2000 is smaller than movies produced after 2000. 

\textbf{Movie Genre Distribution And Performance:}
Generally, each movie can belong to more than three movie genres. For all movies, the top three movie genres on most movies include ``Drama'', ``Adventure'' and ``Fantasy''. Movies belonging to any of those three genres are more than 91\% of the total movies. In order to further analyze the overall genre preference of audiences, we show the violin plot on gross of all movies in Figure \ref{fig:genre}. In this figure, the horizontal bar in each box denotes the median gross of each genre, vertical bar denotes the range of gross in each genre, and the width of the violin shows the quantity of the movie in the same gross.


By comparing the positions of the horizontal bar of each movie genre, we observe that the median movie gross fluctuates widely on different genres. For instance, the median gross of the "Animation" and "Adventure" genres are \$85 million and \$68 million respectively, but those of the ``Film-Noir'' and ``News'' genres only have \$89k and \$95k. Additionally, the box height of some movie genres, like ``News'' and ``Short'', are relatively short compared to the remaining movie genres. By studying the data, we observe that these movie genres are of a relatively small minority, and less than ten movies in total belong to these genres according to our IMDB dataset. 


\subsection{Movie Production Team Statistics}
After analyzing the common movie information, we believe that production team information which influence movie gross are more important than those common movie information. For example, it's more likely that an audience watch a movie due to his/her favorite actress or actor participation. Therefore, in this section, we will analyze some latent movie information such as the movie production team information, which are actor, actress, writer and director. Moreover, we will discuss the movie configuration acquaintance and why it's important to consider it when planning the blockbuster.

\begin{figure*}[htbp]
\centering
\subfigure[Actor]{ \label{fig:Actor}
    \begin{minipage}[l]{.47\columnwidth}
      \centering
      \includegraphics[width=\textwidth]{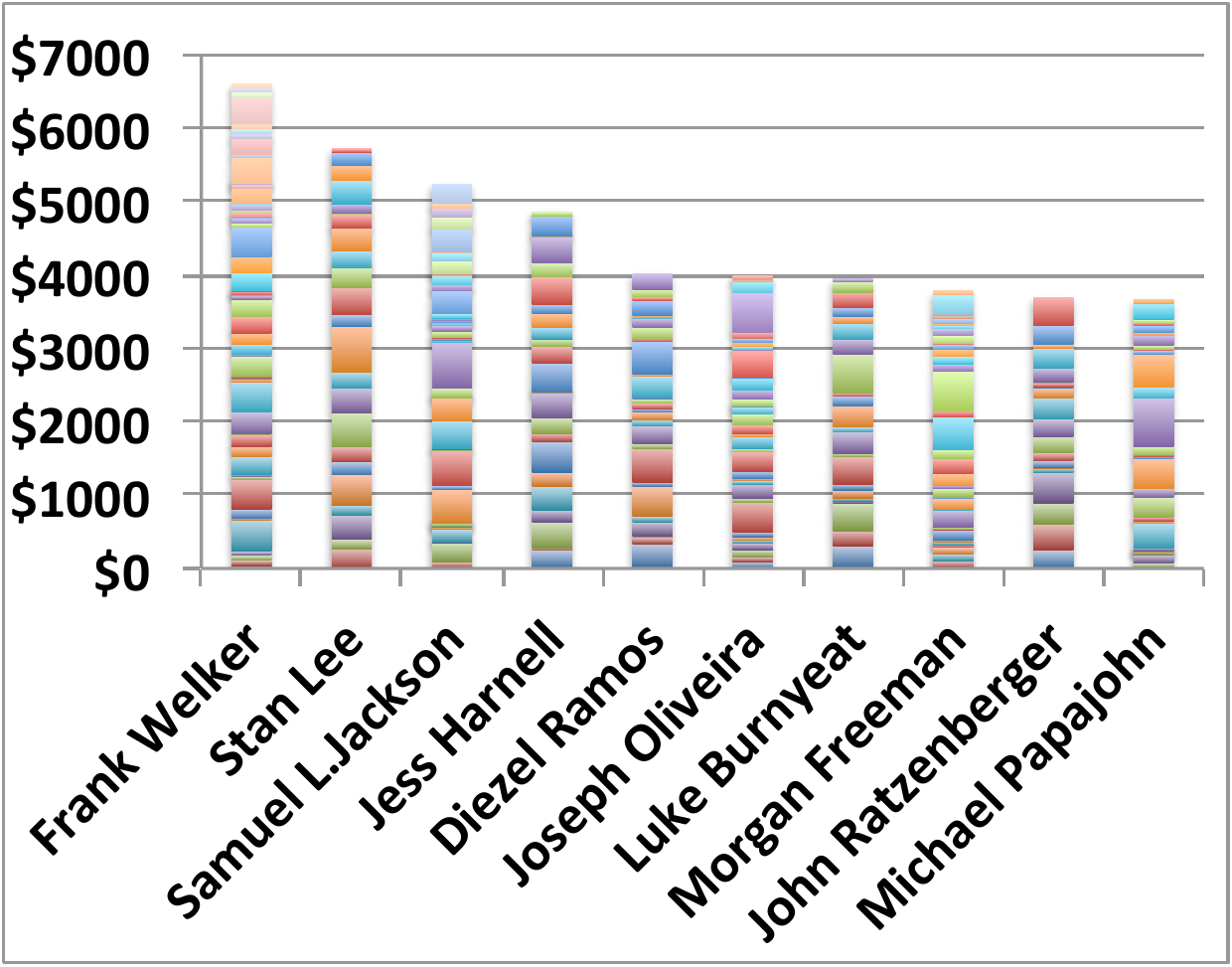}
    \end{minipage}
  }
  \subfigure[Actress]{\label{fig:Actress}
    \begin{minipage}[l]{.47\columnwidth}
      \centering
      \includegraphics[width=\textwidth]{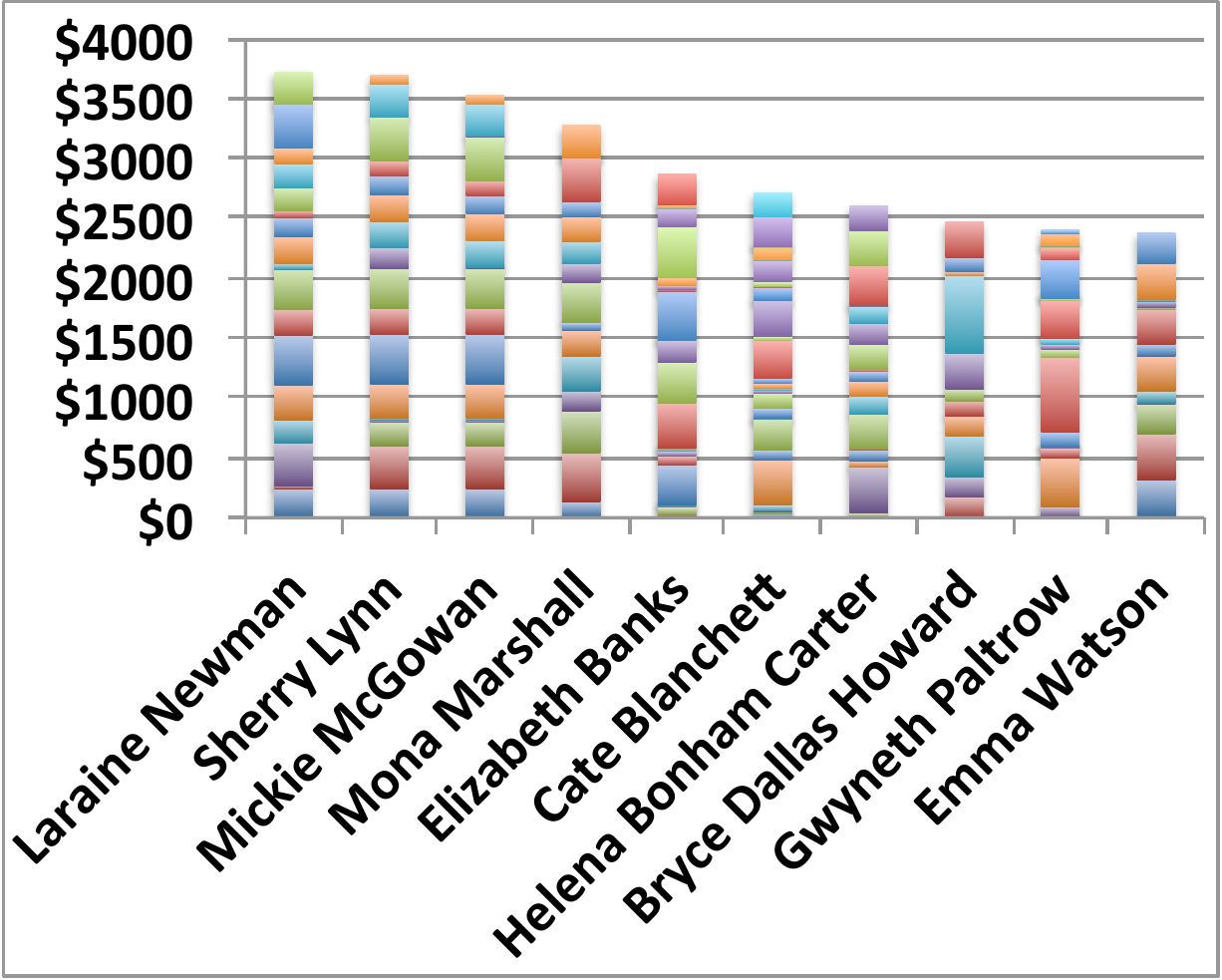}
    \end{minipage}
  }
 \subfigure[Writer]{\label{fig:Writer}
    \begin{minipage}[l]{.47\columnwidth}
      \centering
      \includegraphics[width=\textwidth]{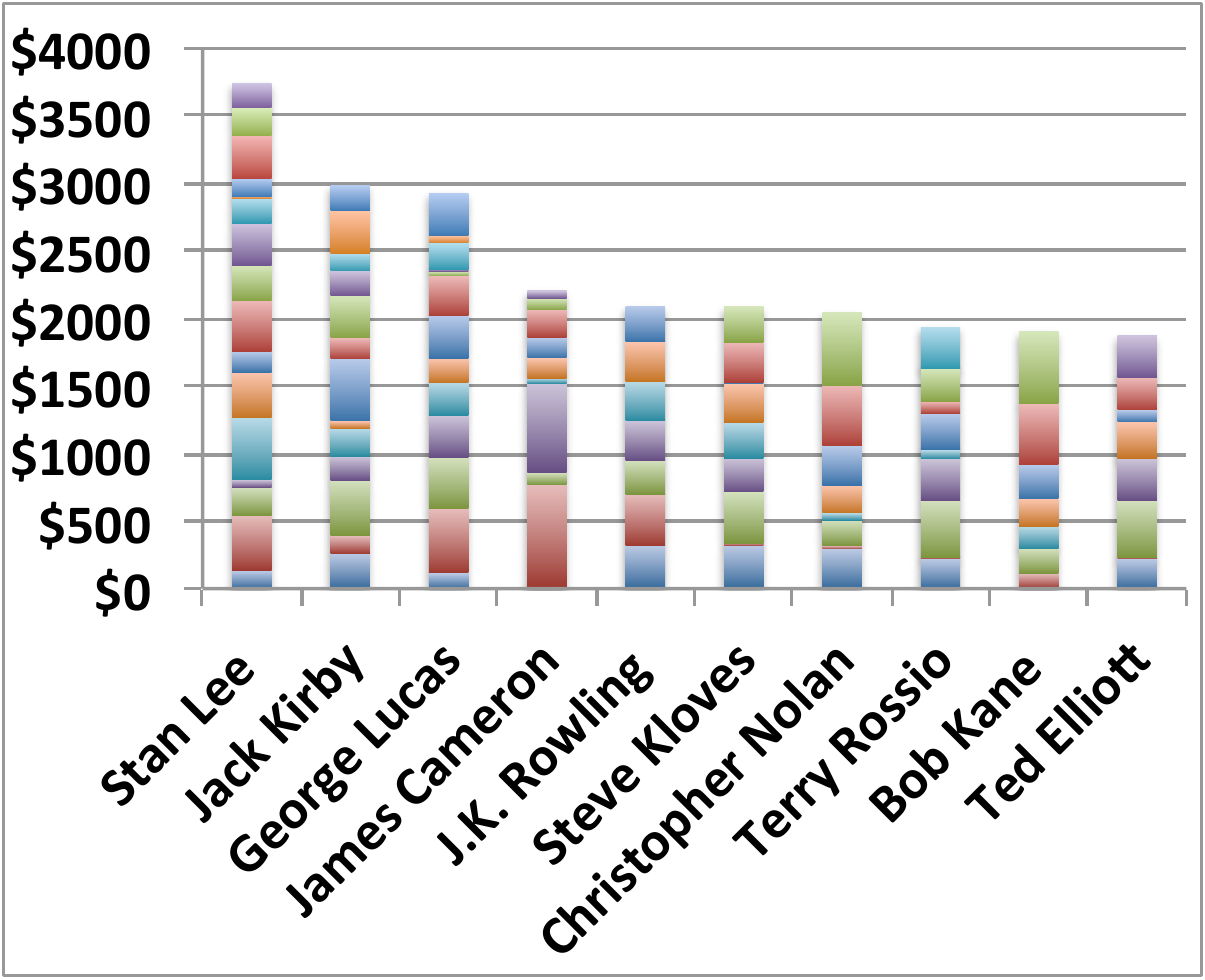}
    \end{minipage}
  }
   \subfigure[Director]{\label{fig:Director}
    \begin{minipage}[l]{.47\columnwidth}
      \centering
      \includegraphics[width=\textwidth]{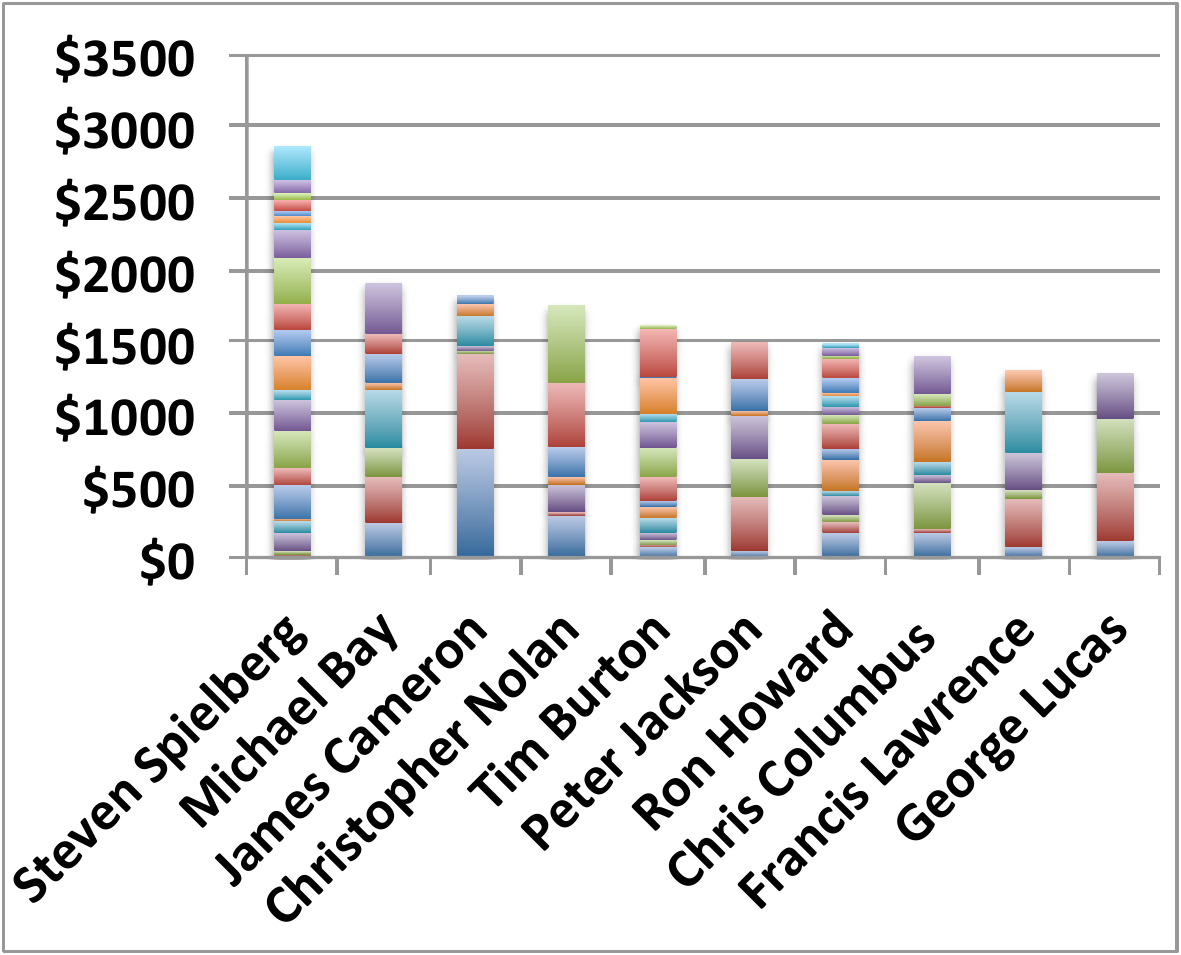}
    \end{minipage}
  }
  \caption{Accumulative Movie Gross vs. Production Team Member. From the left to the right which are Actor, Actress, Writer and Director}
\end{figure*}\label{fig:all}

\begin{table*}[htbp]
  \centering
  \resizebox{7in}{!}{%
    \begin{tabular}{|llcc|llcc|}
    \hline
    \multicolumn{4}{|c|}{\textbf{Actor Actress collaboration}}   & \multicolumn{4}{c|}{\textbf{Actor Writer collaboration}}  \\
    \hline
    \multicolumn{1}{|c}{Actor }&\multicolumn{1}{c}{Actress}&\multicolumn{1}{c}{Times}&\multicolumn{1}{c|}{Genre}&\multicolumn{1}{c}{Actor }&\multicolumn{1}{c}{Writer}&\multicolumn{1}{c}{Times}&\multicolumn{1}{c|}{Genre}  \\
      \hline
    Bernard Lee&Lois Maxwell&7& Action Adventure Thriller& Adam Sandler&Tim Herlihy & 10&Comedy Drama  \\
    Johnny Depp& Helena B. Carter&6&Adventure Fantasy Comedy& Daniel Radcliffe & J.K. Rowling&7&Adventure Family Fantasy \\
    Burt Young&Talia Shire&6&Drama Sport&Bernard Lee & Ian Fleming &7&Action Adventure Thriller\\
    \hline
    \hline 
     \multicolumn{4}{|c|}{\textbf{Actor Director collaboration}}   & \multicolumn{4}{c|}{\textbf{Actress Writer collaboration}}  \\
     \hline
     \multicolumn{1}{|c}{Actor}&\multicolumn{1}{c}{Director}&\multicolumn{1}{c}{Times}&\multicolumn{1}{c|}{Genre}&\multicolumn{1}{c}{Actress }&\multicolumn{1}{c}{Writer}&\multicolumn{1}{c}{Times}&\multicolumn{1}{c|}{Genre}  \\
     \hline
     Johnny Depp&Tim Burton&8&Horror Comedy Drama&Lois Maxwell&Ian Fleming&14&Action Adventure Thriller\\
     Adam Sandler&Dennis Dugan&7&Comedy Romance&Lois Maxwel&Richard Maibaum&11&Action Adventure Sci-Fi\\
     Antonio Banderas&Robert Rodriguez&7&Action Adventure Crime&Mia Farrow&Woody Allen&11&Comedy Drama\\
    \hline
    \hline
     \multicolumn{4}{|c|}{\textbf{Actress Director collaboration}}  & \multicolumn{4}{c|}{\textbf{Writer Director collaboration}}   \\
    \hline
     \multicolumn{1}{|c}{Actress}&\multicolumn{1}{c}{Director}&\multicolumn{1}{c}{Times}&\multicolumn{1}{c|}{Genre}&\multicolumn{1}{c}{Writer }&\multicolumn{1}{c}{Director}&\multicolumn{1}{c}{Times}&\multicolumn{1}{c|}{Genre}  \\
      \hline
     Mia Farrow&Woody Allen&11&Comedy Drama&Ethan Coen&Joel Coen&16&Comedy Crime\\
    Giannina Facio&Ridley Scott&8&Comedy Drama&Fran Walsh&Peter Jackson&10&Adventure Fantasy\\
     Helena B. Carter&Tim Burton&7&Fantasy Adventure Horror&Bobby Farrelly&Peter Farrelly&8&Comedy Romance \\
    \hline
    \end{tabular}
    }
    \caption{The collaboration Among Production Team}
  \label{tab:relation1}%
\end{table*}
\subsubsection{Production Team and Movie Gross}
We show the stacked bar plot of the top ten movie production team members whose movies have the highest accumulative gross. In each stacked bar, the different color represents different movies. The height of the bar represents gross of the given movie, and the higher the bar is, the higher the gross is. 

\textbf{Movie Gross vs. Actor:}
Frank Welker's movies have the highest gross according to Figure \ref{fig:Actor}; He participated in 66 movies based on our dataset and most of them, he acts as voice actor. His movies earned a total of $\$6,579.99$ million with an average of $\$99.7$ million. Among the top ten gross maker actors, Stan Lee is the actor who has the highest average gross of $\$217.8$ million and he is second highest grossing actor. From Figure \ref{fig:Actor}, we can know that most of the those actors act in an average of more than 30 movies, which means that the famous actors are very popular.

\textbf{Movie Gross vs. Actress:}
Compared to actors, actresses relatively act in less movies as shown in Figure \ref{fig:Actress}. Cate Blanchett is the actress who acts in the highest number of movies; She was involved in 35 movies, and she ranks fifth in the top ten actresses. The number of movies she has acted in is much less than Frank Welker. That shows that actors usually act in more movies than actresses. Generally speaking, we can make a conclusion that actresses act in fewer movies than actors.

\textbf{Movie Gross vs. Writer:}
From Figure \ref{fig:Writer}, we observe that the total gross difference between the last seven writers is not obvious. Their movies' gross only have a few million difference. But compared to the first writer and the last writer, their difference is observable. The total gross of first writer is twice that of the last writer, which shows that famous writers are in great demand.

\textbf{Movie Gross vs. Director:}
Directors are responsible for the whole production process of the movies, and their crucial roles may determine the movie quality. The Figure \ref{fig:Director} shows that top ten directors participated in relatively few movies. Steven Spielberg participated in $23$ movies which is the highest of all the directors in our dataset. The average gross of each director is around $\$165.7$ million dollars, which is higher than actor, actress and writer. We can see that best directors relatively act less time of making movie, but each movie they making have a high gross. 

We show the different character between actor, actress, writer and director regarding to movie gross. But they all have a strong relation to the movie gross and are the necessary factors for planning the blockbuster. 

\subsubsection{Movie Configuration Acquaintance} \label{sec:castpref}
We show that the production team has a strong connection with the movie gross. Some of them are the guarantee to a high gross movie. To ensure a high gross movie, effective collaboration among {\cast} also needs to be analyzed. We will see that production members have a strong collaboration with each other. There are six different types of collaborations, which are shown in the Table \ref{tab:relation1}. 

\textbf{Actor and Actress Collaboration:} Bernard Lee and Lois Maxwell participated in seven movies, like ``Dr. No'' which is the first James Bond film, and the genre of those movies are ``Action'', ``Adventure'' and ``Thriller''. The second frequent collaborating partners are Johnny Depp and Helena B. Carter. They participated in recent well-known movies, like ``Sweeney Todd: The Demon Barber of Fleet Street'' and ``Alice in Wonderland''.

\textbf{Actor and Writer Collaboration:} Daniel Radcliffe and J.K. Rowling collaborated in the series of ``Harry Potter''. For the third frequent partners, Bernard Lee and Ian Fleming collaborated in many ``Action'', ``Adventure'' or ``Thriller'' genre movies and actress Lois Maxwell also participated in most of those movies, which shows strong collaboration among the three of them.

\textbf{Actor and Director Collaboration:} Johnny Depp and Tim Burton collaborated in eight movies which makes them the most frequent partners. Among those eight movies, Helena B. Carter also participated in five of them, like ``Corpse Bride'' and ``Dark Shadows''. Adam Sandler and Dennis Dugan collaborated seven times, and in those movies, Tim Herlihy also participated.

\textbf{Actress and Writer Collaboration:} Lois Maxwell was a famous actress during the 1960s and 1970s. She collaborated fourteen times with Ian Fleming and eleven times with Richard Maibaum. The most common genres they participated in are ``Action'', ``Thriller'' and ``Adventure''.

\textbf{Actress and Director Collaboration:} Mia Farrow and Woody Allen collaborated in the ``Comedy'' or ``Drama'' genre movies eleven times and those movies have ``Comedy'' or ``Drama'' genre.

\textbf{Director and Writer Collaboration:} Directors and writers collaborate more often than the other relationships. Ethan Coen and Joel Coen even collaborated sixteen times, with most of the movie genres being ``Comedy'' or ``Crime''.

All of these collaborations show a strong relationship between the production team members and  movie genre. The movies with high collaborations have a high gross and are well-known by viewers. Besides, the binary relationship cannot fully represent those collaborations. For example, Bernard Lee as actor, Ian Fleming as actress and Ian Fleming as writer, participated in many movies together. Moreover, those movies have the same genre, ``Action'', ``Adventure'' and ``Thriller'', which shows that only considering the collaboration between team members is not enough. Instead, considering the collaboration between team members based on movie genre is necessary. In our dataset, there are a lot of same or more complex collaborations like this. Therefore, the movie configuration acquaintance must need be studied when we make the blockbuster planning. The more details of movie configuration acquaintance term will be discussed in the Section \ref{sec:pref}.

\section{MOVIE CONFIGURATION VERIFICATION}\label{sec_ver}
To verify the effectiveness of these factors aforementioned on estimating the movie gross and budget and to learn the weight of movie configuration, in this section, we will build a prediction model to learn their correlations. A set of features (i.e., the configuration) will be extracted for the movies based on each of the factors first. After that, a regression model will be built to project the movie configurations to their budget and gross.

\subsection{Feature Extraction and Movie Budget/Gross Estimation} \label{model}
Features like actor, actress, writer and director are a bag-of-words. Moreover, the relationship between a movie and its feature is one-to-many. For example, each movie belongs to more than one genre. And each movie has more than one actor or actress. 

We use $e$ to represent an element in the movie $m_{i}$ configuration $\textbf{x}_{(m_{i})}$=$[\textbf{x}_{(m_{i})}^t,~\textbf{x}_{(m_{i})}^s,~\textbf{x}_{(m_{i})}^d,~\textbf{x}_{(m_{i})}^w,~\textbf{x}_{(m_{i})}^g]$. We use the binary value to set the element. Namely, for example, if actor $t_j$ participates in movie $m_i$, then $\textbf{x}_{(m_{i})}^{t_{j}}$ equals to 1, otherwise, it equals to 0. In the same way, we can get the vector representation of $\textbf{x}_{(m_{i})}^s$, $\textbf{x}_{(m_{i})}^d$, $\textbf{x}_{(m_{i})}^w$ and $\textbf{x}_{(m_{i})}^g$.

After extracting all the features of a movie, we can train an approximation model of budget function and gross function. Since the cost and income of each production team member can not be negative, we use Lasso linear regression \cite{tibshirani1996regression} and force the coefficients to be non-negative. Formally, they can be represented as:
\begin{align}
Budget(\textbf{x}) = \min_{\textbf{w}_{b},b_{b}}||B - (\textbf{w}_{b}^T\textbf{x}+b_{b})||_{2}^2 + \lambda ||\textbf{w}_{b}||_{1}
~s.t. \textbf{w}_{b} \geq 0 
\end{align}
\begin{align}
Gross(\textbf{x}) = \min_{\textbf{w}_{g},b_{g}}||G - (\textbf{w}_{g}^T[B,\textbf{x}]+b_{g})||_{2}^2 + \lambda||\textbf{w}_{g}||_{1}
~~ s.t. \textbf{w}_{g} \geq 0 
\end{align}
where $\textbf{x}$ is the movie configuration and $B$ is the budget and $G$ is the gross of movie in the IMDB knowledge base. $\textbf{w}_{b}$ and $\textbf{w}_{g}$ are the weights and $b_{b}$ and $b_{g}$ are the intercepts of function $Budget(\textbf{x})$ and function $Gross(\textbf{x})$ function respectively. $\lambda$ is the coefficient of the $\ell_{1}$-norm regularization, which we set to 0.1 in the experiment.

\subsection{Movie Budget/Gross Estimation Experiment Results} \label{estimation}
In the experiments, 80\% of $3156$ movies are used as training data and 20\% are used as testing data. The 5-fold cross validations are performed on training data. We analyze the effectiveness of those features on function $Budget(\textbf{x})$ and function $Gross(\textbf{x})$ separately. 
To measure the performance, we use the mean absolute percentage error (MAPE) as the evaluation metrics which represent as follow:
\begin{align}
MAPE= \frac{100}{n} * \sum_{t=1}^{n}\left|\frac{(A_{t}-E_t)} {A_t}\right|
\end{align}
where $A_t$ is the actual value and $E_t$ is the estimated value.

The following is our compared methods:

\textbf{Models using all information}
\begin{itemize}
    \item ALL: Method ALL builds the gross approximation with all features, which are genre, actor, actress, writer and director.
\end{itemize}

\textbf{Models using partial information}
\begin{itemize}
    \item Genre: Method Genre builds the gross and budget approximation model with the genre information alone.
    \item Actor: Method Actor just uses the actor feature to build the gross and budget approximation model.
    \item Actress: Method Actress builds the gross and budget approximation model with the actress feature.
    \item Writer: Method Writer builds the gross and budget approximation model with writer feature.
    \item Director: Method Director builds the gross and budget approximation model with only director feature.
\end{itemize}
\begin{figure}[t]
\includegraphics[width=8.5cm]{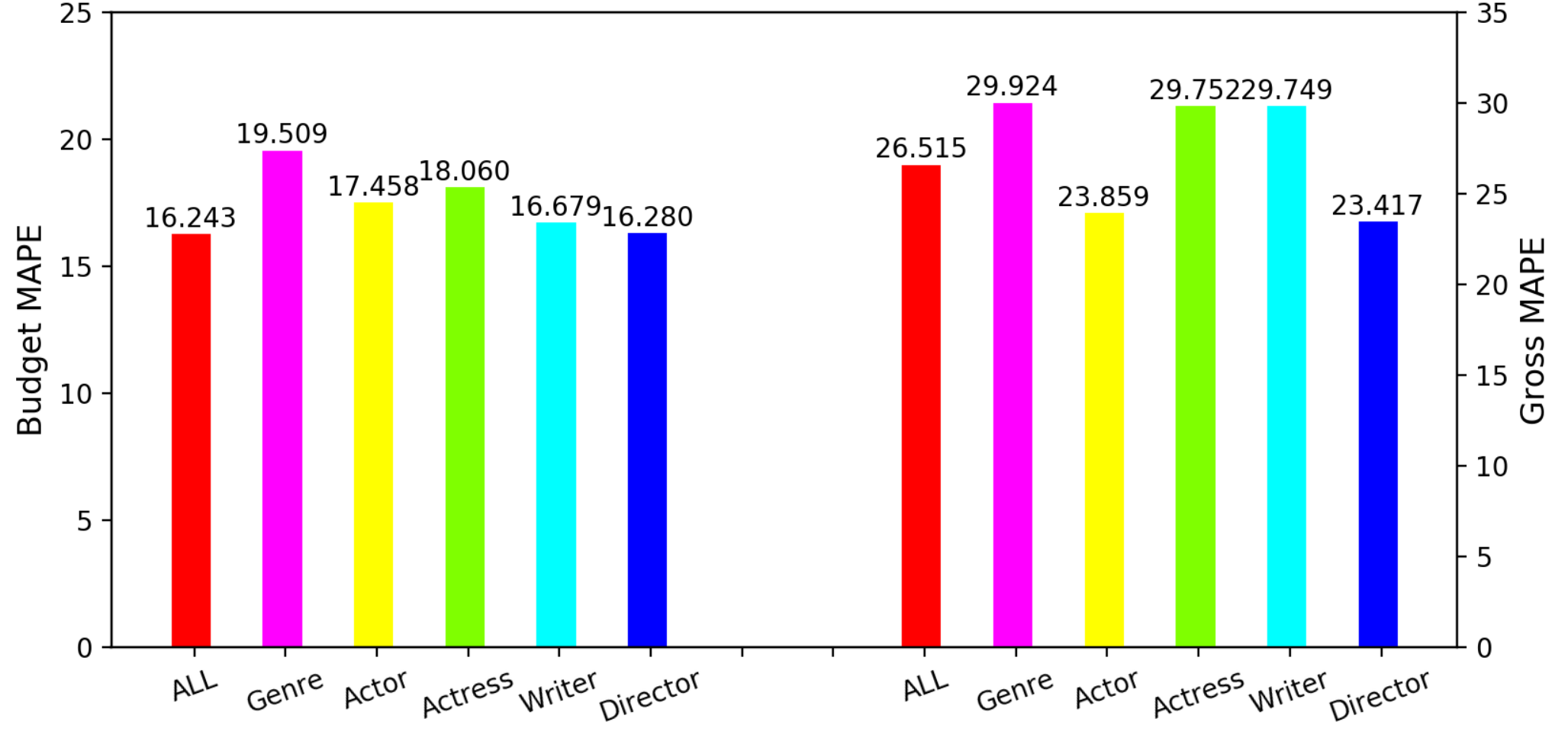}
\caption{The MAPE of different approximation models}
\label{fig:acc}
\end{figure}

The experiment results are available in Figure \ref{fig:acc}. The left part shows the MAPE on predicting the movie budget and the right part shows the MAPE of movie gross estimated by different methods. 

Using all features together gives us the lowest MAPE, 16.243$\%$ as shown in the left part of Figure \ref{fig:acc}. When looking at the production team information, we find that writer and director have relative low MAPE, $16.679\%$ and $16.280\%$, which implies writer and director positively correlated to the movie budget. While in reality, the salary of writer and director can determine how much the movie producer needs to invest in the movie. Production team information achieves a relatively lower MAPE than genre, due to wide difference of movie budgets in the same genre; therefore genre is not a good factor for the budget. 

By comparing all features in the gross approximation, we can observe that Director achieves the lowest MAPE (23.417$\%$). Actor gets the second-lowest MAPE (23.859$\%$). Compared to those two models, the performance of ALL is not good. It's probably because feature actress, writer and genre have a large MAPE, meaning that those features have no (or weak) correlation to the movie gross. Therefore, combining all features together will achieve 26.515$\%$ on MAPE. Such result is reasonable because a director with great reputation is more likely to produce a good movie. Moreover, similarly as the conclusion in Section 4, production team information can reach a relatively lower MAPE than the genre.

Even if the performance of ALL is not as good as same only feature such as director or actor, the performance is still good. And our goal is to learn a function which can map the movie configuration with gross and budget. 

\section{BLOCKBUSTER PLANNING: BigMovie}\label{sec_plan}
Since we have already demonstrated that good collaboration between production team members is the safeguard for the profit of movie. In this section, we first formulate the movie configuration acquaintance. Based on the movie gross, budget estimation and movie configuration acquaintance function, we provide the joint objective function of {\ours} and a cubic programming algorithm to effectively solve the objective.

\subsection{Movie Configuration Acquaintance}\label{sec:pref}
As we discussed previously, it is advantageous to have production team members that have a strong collaboration to the other specific members and furthermore have strong acquaintance to the certain movie genre. If team members have already participated together before, they will have chemistry when they participate in the next movie, which may stimulate the increase of movie gross. Besides, production team members that have joined in a certain movie genre previously can more easily work together when making the same genre type movies. Those two types of acquaintances have a great effect on the movie gross and movie budget, so finding the mathematical representation of the movie configuration acquaintance is important.  

\textbf{Movie Configuration Acquaintance Function}:
We discuss in section \ref{sec:castpref} the binary relationship between two members cannot well represent their collaboration and movie genre need to be considered as well. To solve these problems, we use a three dimensional tensor $\textbf{W}_{a}\in \mathbb R^{C\times C \times G}$ to represent movie configuration acquaintance, where the $C$ is the dimension for the size of all cast, $G$ is the dimension for the size of all movie genres. 
We propose to define the movie configuration acquaintance as follow:
\begin{equation}
Acquaintance(\textbf{x})=\sum_{n=0}^{C-1}\sum_{m=0}^{C-1}\sum_{l=0}^{G-1}\textbf{W}_{a}[n][m][l]\cdot\textbf{x}[n]\cdot\textbf{x}[m]\cdot\textbf{x}[l],
\end{equation} 
where $n$ and $m$ are the production team member $\in \mathcal{C}$. And $l$ is the movie genre $\in \mathcal{A}_{(m_{i})}^g$
\subsection{Joint Objective Function} \label{sec:object}
Maximize movie gross can be mathematically represented as:
$
\max_{\textbf{x}}\sum_{i=0}^{N-1} 
\textbf{w}_{g}[i+1] \cdot \textbf{x}[i]+b_{g}+\textbf{w}_{g}[0]\cdot B,
$
where $\textbf{w}_{g}$ is the weight and $b_{g}$ is the intercept of function $Gross(\textbf{x})$ that we learned from section \ref{model}.
The movie budget bound can be mathematically represented as:
$
    \sum_{i=0}^{N-1}\textbf{w}_{b}[i]\cdot\textbf{x}[i]+b_{b}\leq B,
$
where $\textbf{w}_{b}$ and $b_{b}$ are weights and intercepts of function $Budget(\textbf{x})$ we learned from section \ref{model}.

The objective of the BP problem is to find the optimal movie configuration that can maximize the movie gross and movie configuration acquaintance while not exceeding the movie budget bound. So the joint objective function represents as:
\begin{align}\label{eq:objt}
& \max_{\textbf{x}}\alpha (\sum_{i=0}^{N-1} 
\textbf{w}_{g}[i+1] \cdot \textbf{x}[i]+b_{g}+\textbf{w}_{g}[0]\cdot B) \\ \nonumber
&+\beta \sum_{n=0}^{C-1}\sum_{m=0}^{C-1}\sum_{l=0}^{G-1} \textbf{W}_{a}[n][m][l]\cdot\textbf{x}[n]\cdot\textbf{x}[m]\cdot\textbf{x}[l]\\\nonumber
&s.t. \sum_{i=0}^{N-1}\textbf{w}_{b}[i]\cdot\textbf{x}[i]+b_{b}\leq B, \qquad \forall i: \ \textbf{x}_{i} \in \{0,1\} \nonumber
\end{align}
where $\alpha$ and $\beta$ are parameters to adjust movie gross estimation and movie configuration acquaintance which are studied in Section \ref{sec_exp}.
\begin{figure*}[t]
\centering
  \subfigure[Accuracy with different $\beta$]{\label{fig:c_f1_alpha}
    \begin{minipage}[l]{.47\columnwidth}
      \centering
      \includegraphics[width=\textwidth]{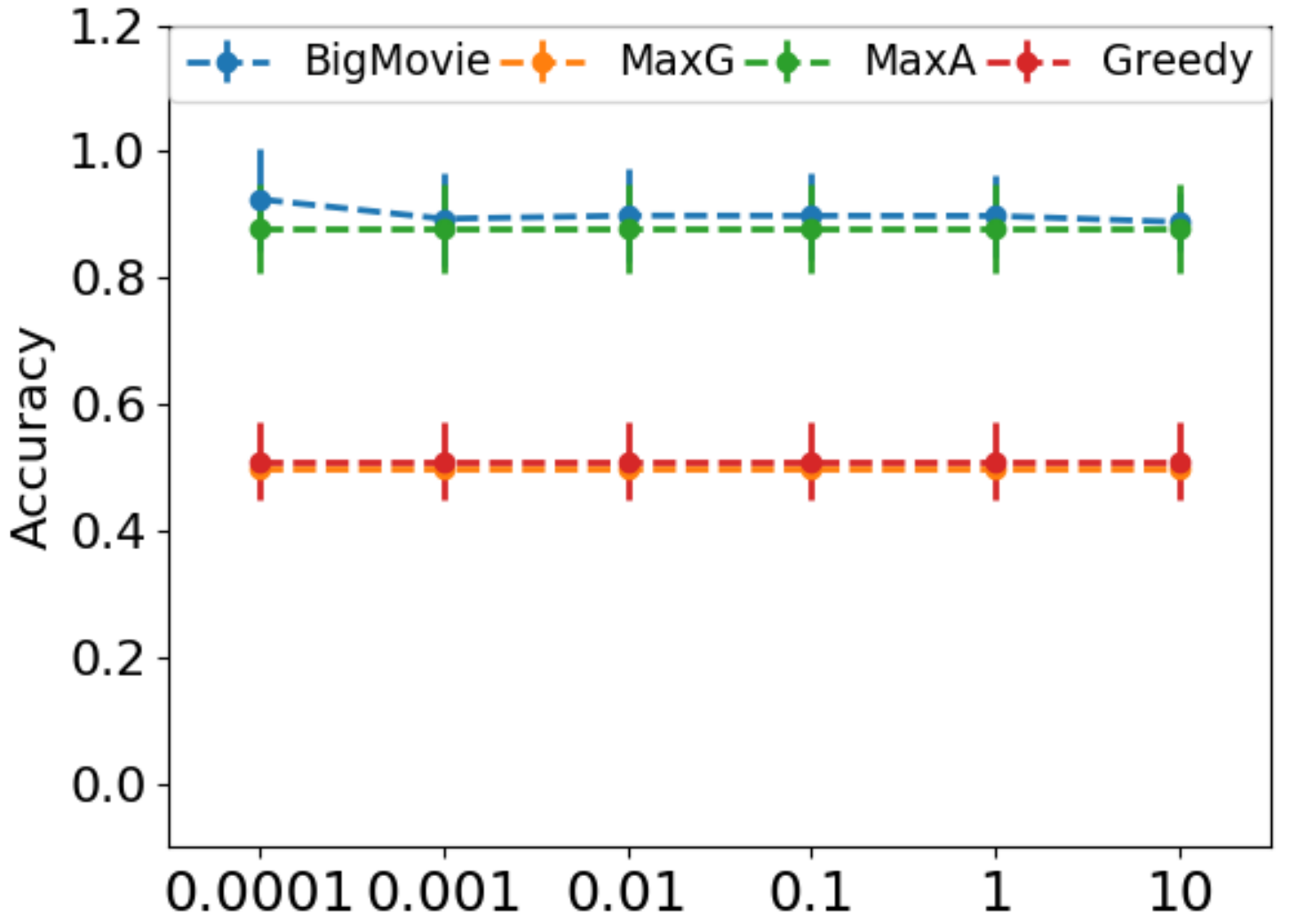}
    \end{minipage}
  }
\subfigure[$F1$ with different $\beta$]{ \label{fig:c_acc_alpha}
    \begin{minipage}[l]{.47\columnwidth}
      \centering
      \includegraphics[width=\textwidth]{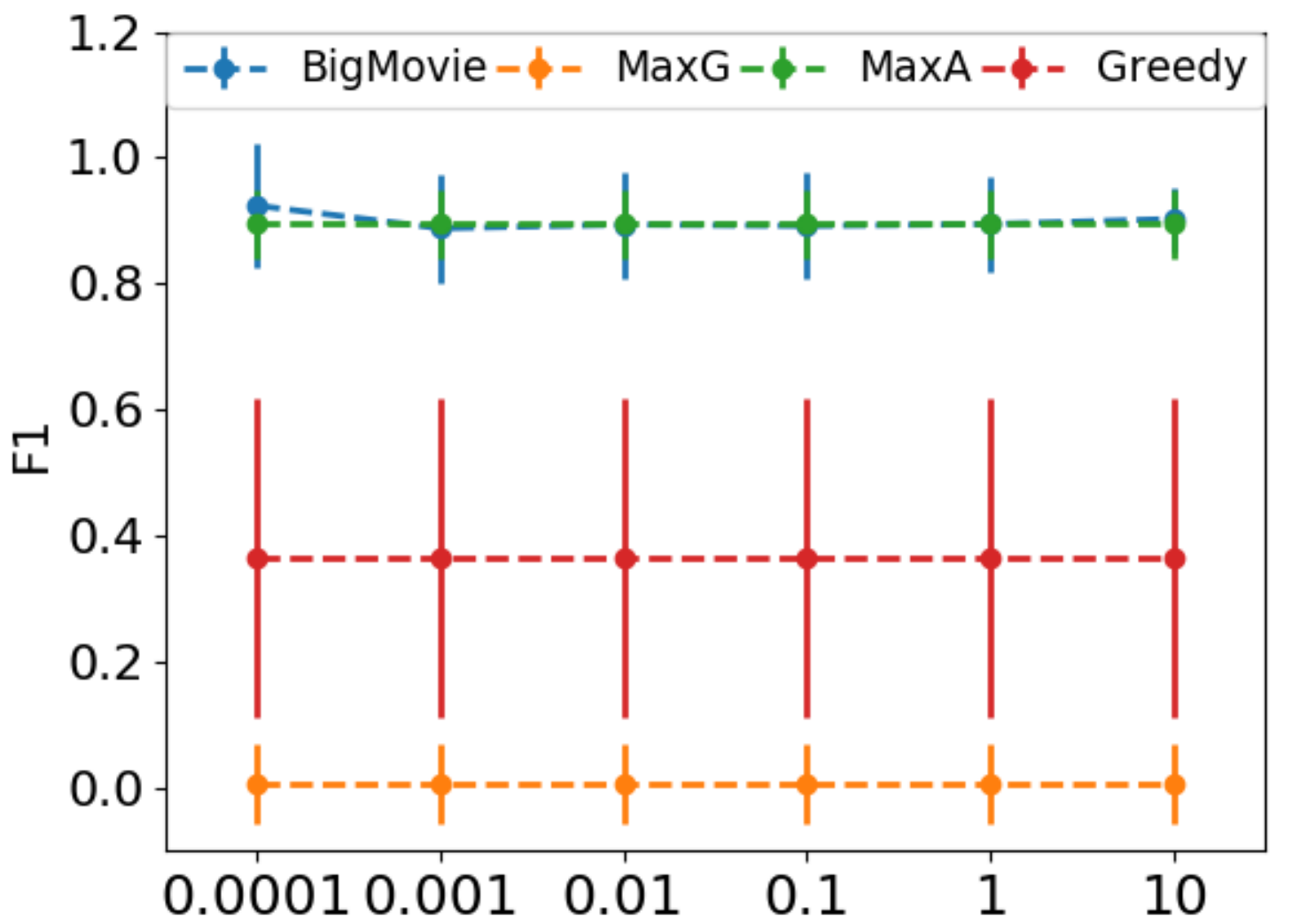}
    \end{minipage}
  }
  \subfigure[Accuracy with different predicted ratio]{\label{fig:c_f1_p}
    \begin{minipage}[l]{.47\columnwidth}
      \centering
      \includegraphics[width=\textwidth]{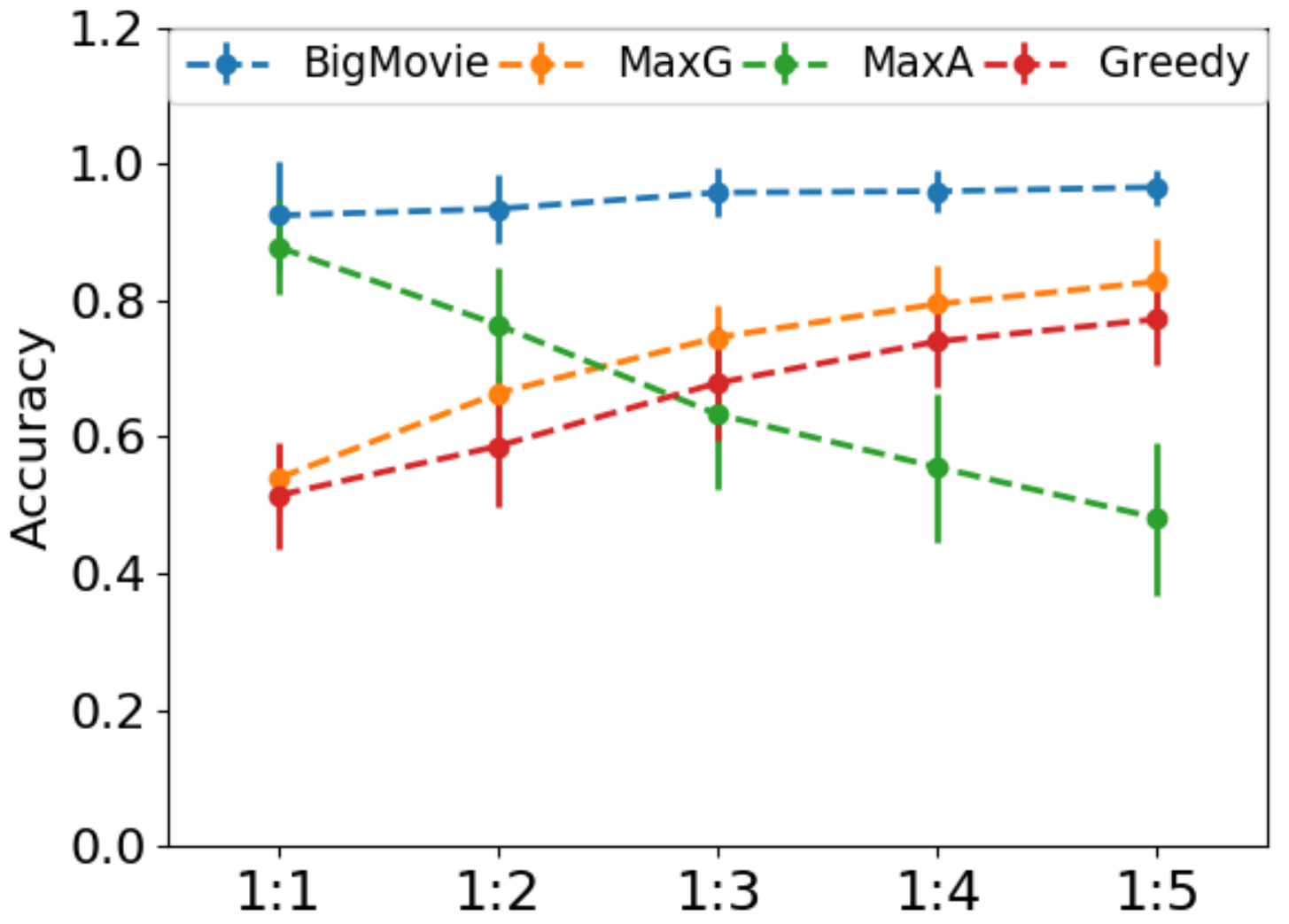}
    \end{minipage}
  }
   \subfigure[$F1$ with different predicted ratio]{\label{fig:c_acc_p}
    \begin{minipage}[l]{.47\columnwidth}
      \centering
      \includegraphics[width=\textwidth]{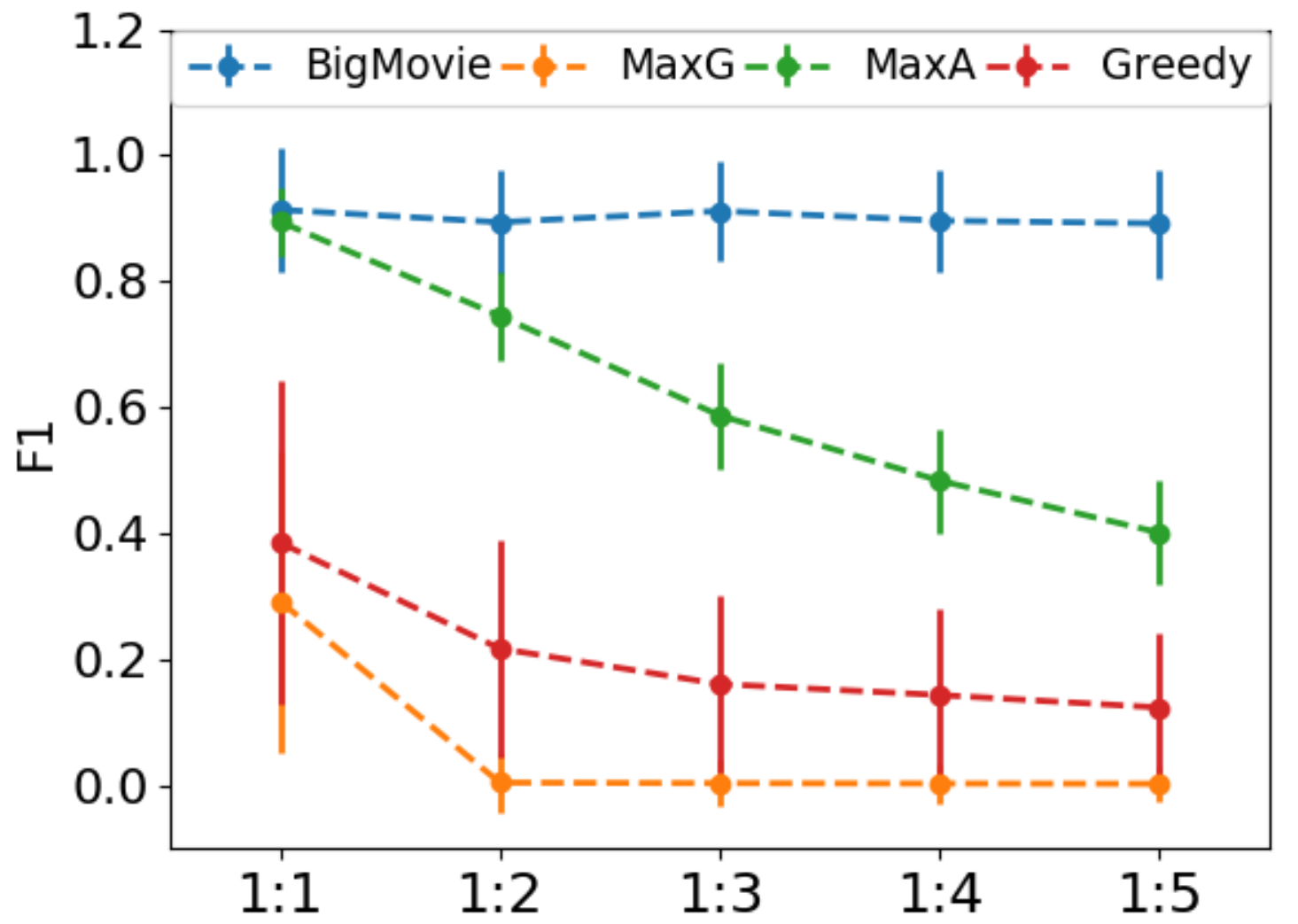}
    \end{minipage}
  }
  \caption{Planning production team with different $\beta$ and different predicted ratio measured by Accuracy and $F1$}
\end{figure*}

\begin{figure*}[t]
\centering
  \subfigure[Accuracy with different $\beta$]{\label{fig:g_f1_alpha}
    \begin{minipage}[l]{.47\columnwidth}
      \centering
      \includegraphics[width=\textwidth]{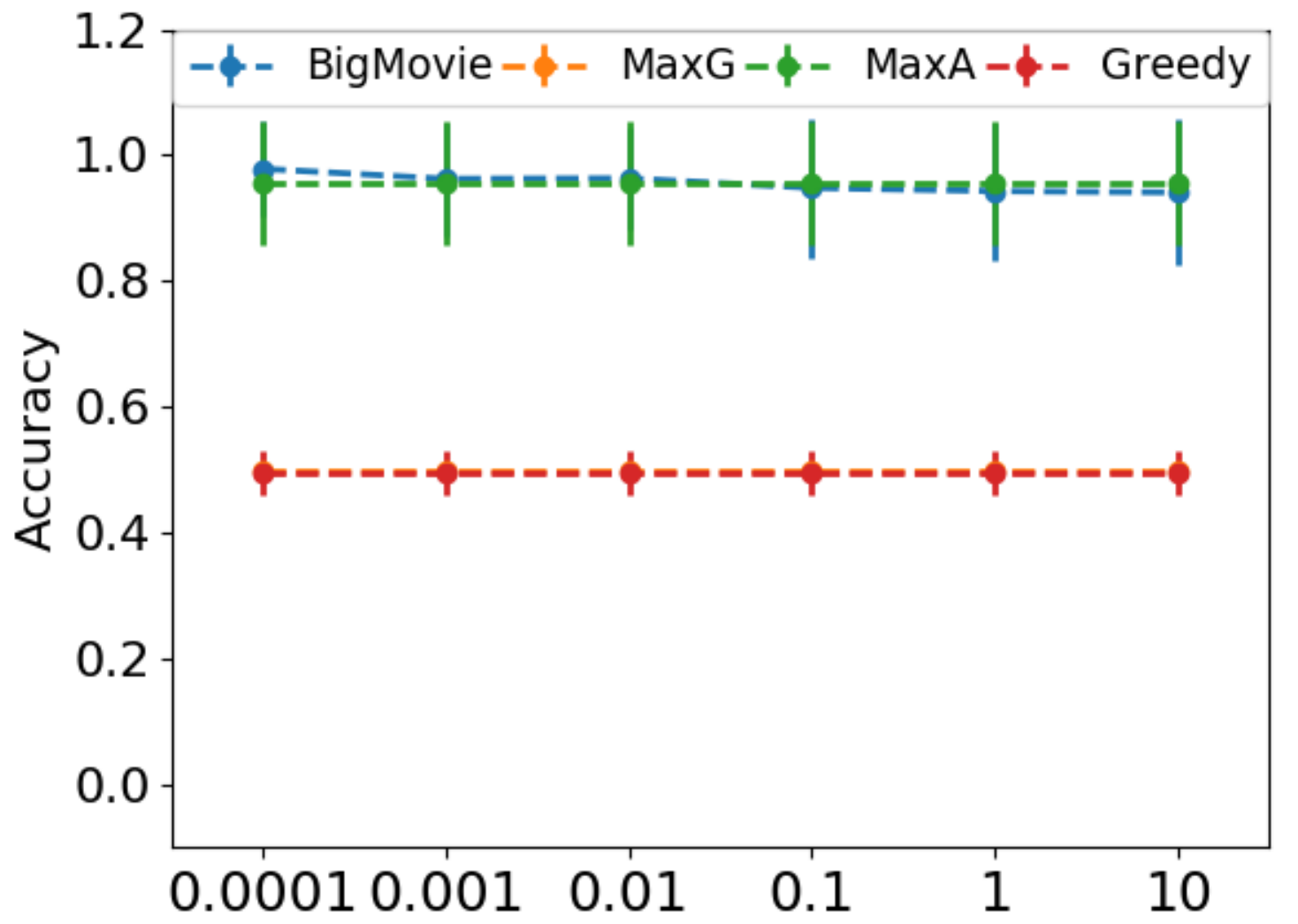}
    \end{minipage}
  }
  \subfigure[$F1$ with different $\beta$]{ \label{fig:g_acc_alpha}
    \begin{minipage}[l]{.47\columnwidth}
      \centering
      \includegraphics[width=\textwidth]{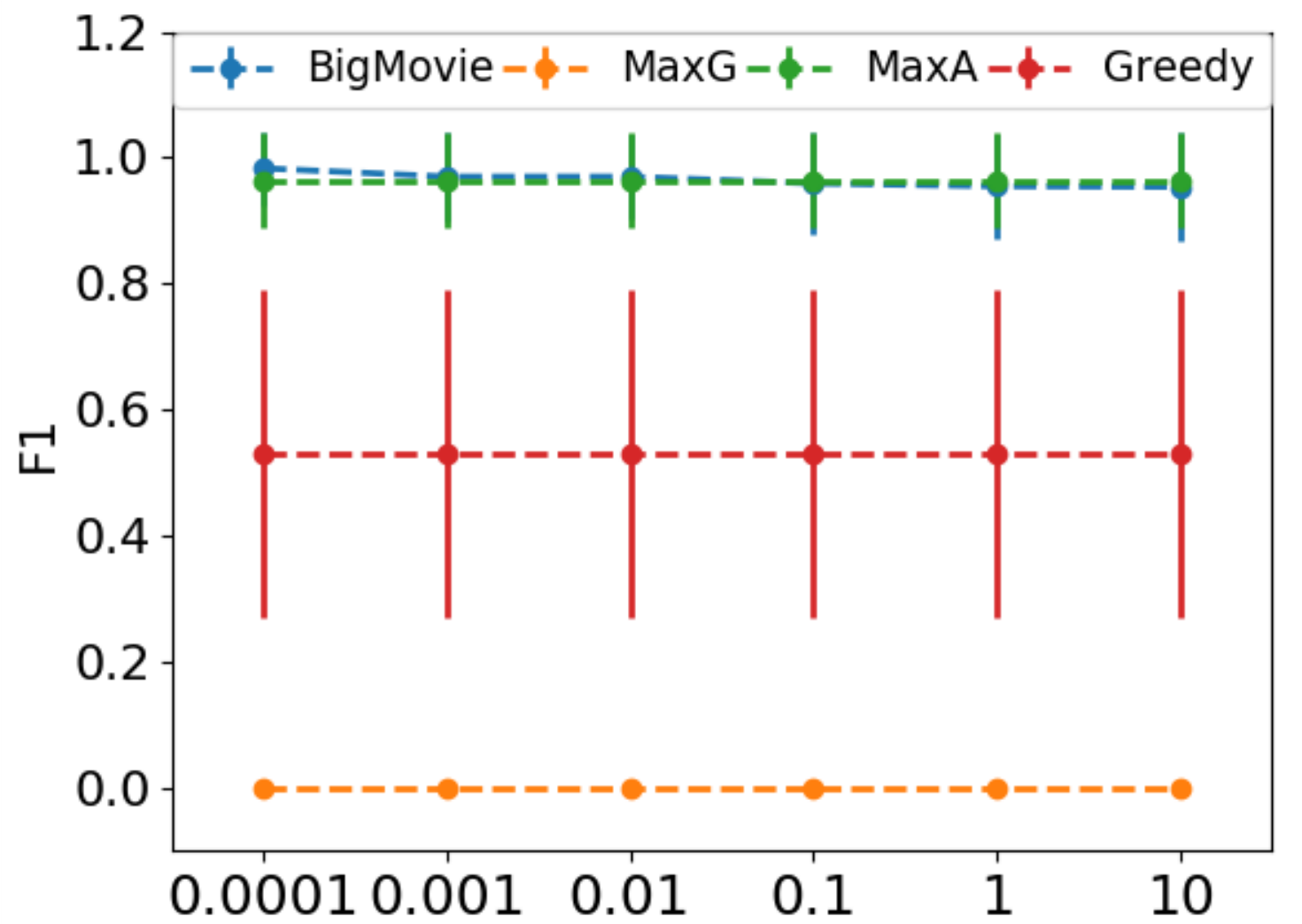}
    \end{minipage}
  }
\subfigure[Accuracy with different predicted ratio]{\label{fig:g_f1_p}
    \begin{minipage}[l]{.47\columnwidth}
      \centering
      \includegraphics[width=\textwidth]{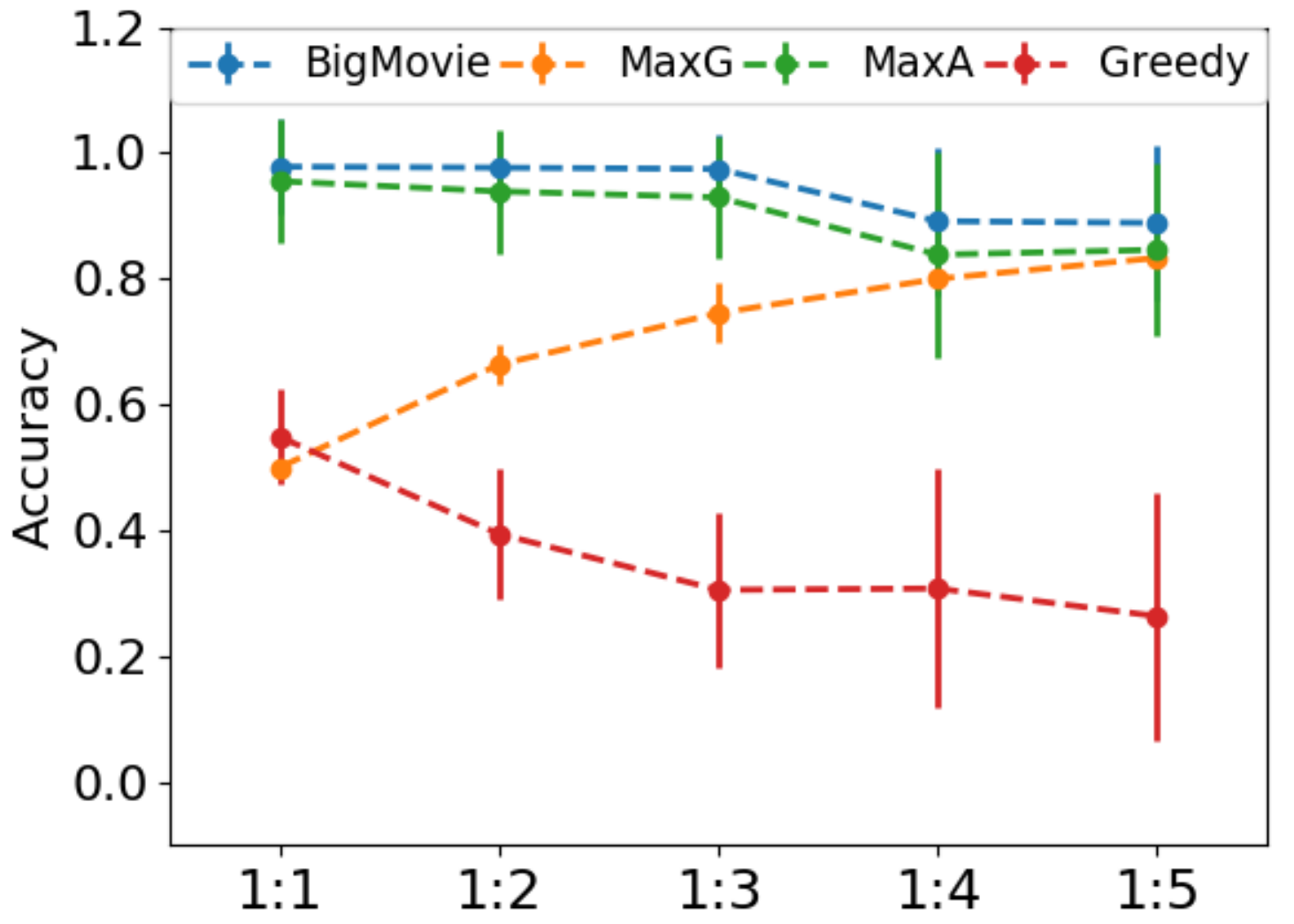}
    \end{minipage}
  }
 \subfigure[$F1$ with different predicted ratio]{\label{fig:g_acc_p}
    \begin{minipage}[l]{.47\columnwidth}
      \centering
      \includegraphics[width=\textwidth]{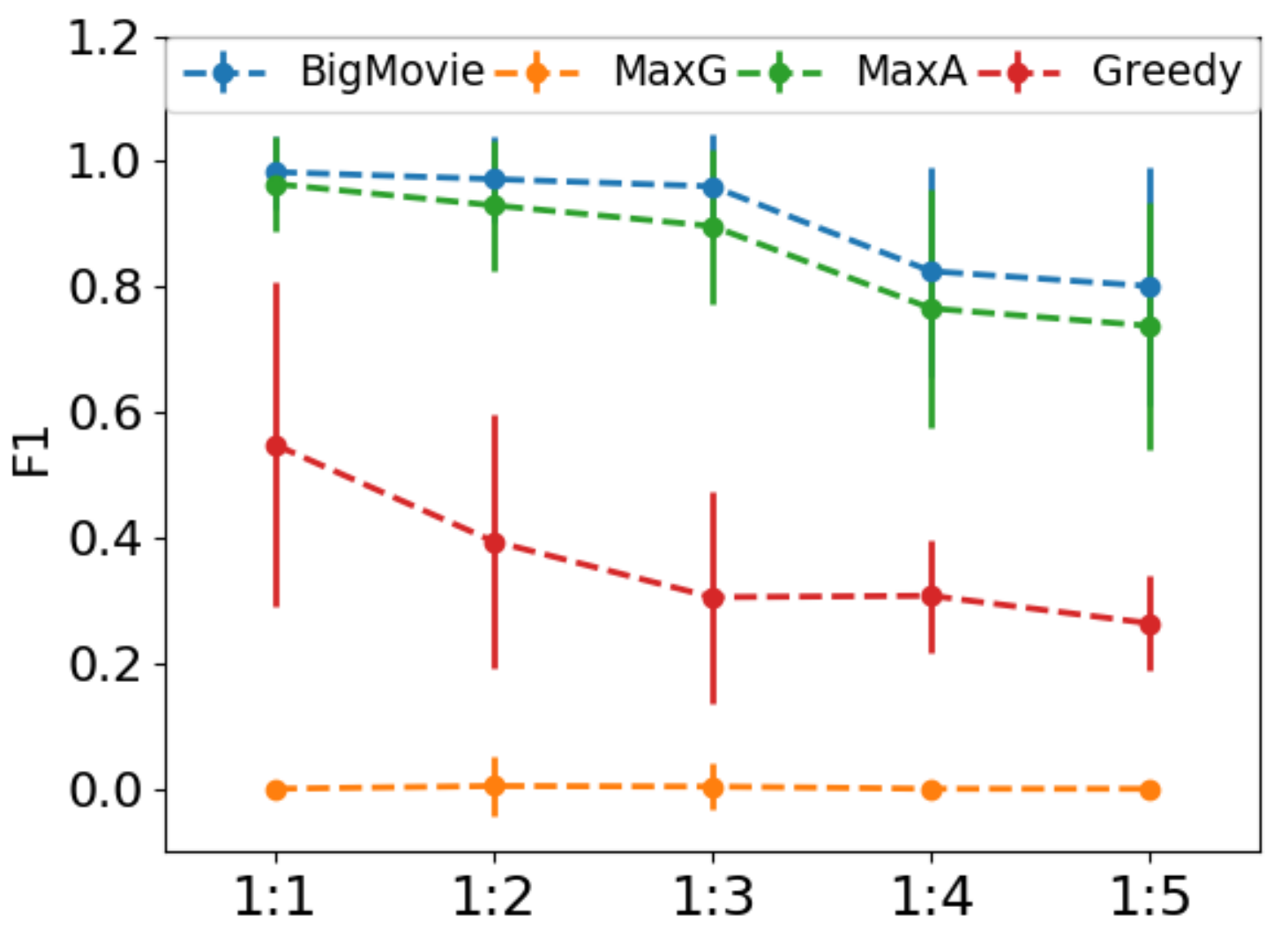}
    \end{minipage}
  }
  \caption{Planning movie genre with different $\beta$ and different predicted ratio measured by Accuracy and $F1$}
\end{figure*}
\subsection{Prove NP-Hardness}
In this section, we prove that the Blockbuster Planning with maximize movie configuration acquaintance problem is a NP-hard problem.
In Equation \ref{eq:objt} of the BP problem, two objectives equations are involved: the gross equation weighted by $\alpha$, and the acquaintance equation weighted by $\beta$. 
By assigning $\alpha = 1$ and $\beta = 0$, we will show that the \textit{Knapsack problem} can be reduced to the BP problem, which is a classic NP-hard problem. Meanwhile, by assigning $\alpha = 0$ and $\beta = 1$, we will show that the \textit{Maximal Clique problem} can be reduced to the BP problem. 

Given a set of items, each with a weight and a value, the \textit{Knapsack problem} aims at picking the items to be included in a bag so that the total weight is less than a given limit while maximizing the total value. By treating items as features in the movie configuration vector $\mathbf{x}$ with corresponding values in vector $\mathbf{w}_g$ and weights in vector $\mathbf{w}_b$, \textit{Knapsack problem} can be exactly reduced to the BP problem (with $\alpha = 1$ and $\beta = 0$), where the bag limit is denoted as the provided budget $B$. If we can identify an optimal movie configuration vector $\mathbf{x}$, the items corresponding to the features with value $1$ can be selected, which will be the optimal solution to the \textit{Knapsack problem}.

Given an undirected graph formed by a finite set of nodes and a set of undirected edges, the \textit{k-Clique problem} aims at determining whether there exist a clique involving $k$ nodes in graph or not. Let a tensor $\mathbf{W}_{p}$ denote whether nodes can form a triangle in the input graph or not. If nodes $n_i, n_j, n_k$ can form a triangle, then $W_{p}[i][j][k] = 1$, and $0$ if not. By treating each node in the graph as a feature to be determined in the movie movie configuration vector $\mathbf{x}$ and assign vectors $\mathbf{w}_g$, $\mathbf{w}_b$ to be $\mathbf{1}$, the problem of identifying a clique of size $k$ in the input graph (i.e., \textit{k-Clique problem}) can be reduced to the problem of obtaining the optimal value of $k(k-1)(k-1)$ in the BP problem, where the budget $B$ takes value $k$. If we can identify an optimal value $k(k-1)(k-1)$, then the nodes corresponding to features with value $1$ will be selected to for a clique of size $k$ in the input graph.

Therefore, the BP problem containing these two objectives  makes itself at least as difficult as the \textit{Knapsack problem} \cite{sahni1975approximate} and the \textit{Maximal Clique problem} \cite{bomze1999maximum}, which renders the BP problem to be NP-hard.

\subsection{Solving the Objective}
Since this problem is NP-complete \cite{papadimitriou1981complexity} and no polynomial-time solutions can solve the problem efficiently, we propose to solve the problem with two steps: (1) integer constraint relaxation, and (2) result post-processing.
We relax the integer constraint on variables, and allow them to take real values in the range of $[0, 1]$ to help address the problem in polynomial time \cite{hochbaum1993tight}. Based on the obtained real-valued solution $\textbf{x}$ denoting the score of the feature-gross links, we post-process the variable to binary values by pruning with a confidence threshold $\theta$ in $[0, 1]$. For the variables, e.g., if $x_{i} > \theta$, we will map $x_{i}$ to value 1; otherwise, $x_{i}$ will be mapped to value 0. After post-processing, we can obtain the final optimal movie plan by finding corresponding genre $g_i$ and team member $c_i$ for the movie $m_i$ having values $x_{i}=1$.

\subsection{Experiments}\label{sec_exp}

In this section, we give the experimental analysis of BigMovie, and evaluate its performance for designing new movies. We seek to answer two main questions:\\
\textbf{Q1.} How well does BigMovie quantitative performance? \\
\textbf{Q2.} Is the movie planned by BigMovie reasonable? \\
\subsubsection{\textbf{A1}: Quantitative Evaluation}

\noindent
\quad In order to quantitatively measure the performance, we use our method to design the movie production team setting or movie genre to see whether our model can match with the ground truth, the movie setting in the database. The {\problem} problem is a new problem, and no existing methods can be applied to address it directly. To show the advantages of the framework {\ours}, we compare some other methods with {\ours} measured by accuracy and F1 score on production team and genre. The comparison methods used in the experiments are listed as follows:

\begin{itemize}
    \item $\textbf{{\ours}}$: framework in the paper that achieves the maximum gross and maximizes the movie configuration acquaintance. By setting $\alpha=1$, then the performance of different $\beta$ will be studied in the experiments.
    \item $\textbf{MaxG}$: aims to maximize the movie gross, which sets the objective function Eq. (\ref{eq:objt}) with $\alpha=1$ and $\beta=0$.
    \item $\textbf{MaxA}$: only considers the movie configuration acquaintance by setting the objective function Eq. (\ref{eq:objt}) with $\alpha=0$ and $\beta=1$.
    \item $\textbf{Greedy}$\cite{dantzig1957discrete}: iteratively picks the greedy choice to maximize movie gross, by always choosing the maximum ratio of $\frac{\textbf{w}_g}{\textbf{w}_b}$.
\end{itemize}

We verify the effectiveness of {\cast} and genre planning on $3,156$ IMDB movies separately. When studying the {\cast} feature, we set the {\cast} as unknown while the movie genre is given, and vice versa.

\begin{tcolorbox}
\textbf{Observation 1.} As shown in Figures 5 and 6, BigMovie outperforms the competing baselines $\textbf{MaxG}$, $\textbf{MaxA}$ and $\textbf{Greedy}$ and obtains higher accuracy and F1 score.
\end{tcolorbox}

From the experiment results, we can see that {\ours} can get more than 90\% accuracy on both genre and production team, which is consistently better than other methods on different $\beta$. When $\beta=0.0001$, the best performance is achieved on production team and genre, as shown in Figures \ref{fig:c_f1_alpha}, \ref{fig:c_acc_alpha} and \ref{fig:g_f1_alpha}, \ref{fig:g_acc_alpha}. With $\beta=0.0001$, we study different ratios of positive and negative samples that are randomly selected, as shown in Figure \ref{fig:c_f1_p}, \ref{fig:c_acc_p}, \ref{fig:g_f1_p} and \ref{fig:g_acc_p}. 

\begin{tcolorbox}
\textbf{Observation 2.} Maximizing both the movie gross and movie configuration acquaintance simultaneously can achieve best performance.
\end{tcolorbox}

For methods only depending on maximizing the gross, MaxG and Greedy, because they do not depend on $\beta$, their performances do not change with different $\beta$. Since we use Lasso with non-negative constraint to learn the weight of budget and gross, most of weights we learned are zeros. Therefore, most candidates are not selected in the production team study, when using MaxG. Such planning is less ideal because not choosing any candidate is not the goal of an optimal planning. The Greedy method achieves higher accuracy than MaxG, but still has the same problem as MaxG. In the genre study, MaxG select all the candidate genre, so when the planning scale gets larger, its performance gets worse. 

For the MaxA method which only uses movie configuration acquaintance part of objective function, in both production team and genre planning, MaxA doesn't have the problem of not selecting any samples. This shows the importance of considering movie configuration acquaintance. But when the negative ratio gets bigger, MaxA performs worse. Since {\ours} outperforms MaxA, the results show the importance of considering simultaneously maximizing the movie gross and movie configuration acquaintance.   

\subsubsection{\textbf{A2}: Case Study}
\ 

\noindent
\quad We show a case study to demonstrate the reasonable and effectiveness of the proposed method. We choose ``The Avengers'' to plan, which has the fourth-highest movie gross on our dataset, \$623.27 million. For fairness, we delete the other sequel movies of ``Avengers''. We provide movie genre ``Action'', ``Adventure'' and ``Sci-Fi'', movie budget and 250 random candidates to build an about 20 casts production team. The detail of planned movie configuration and actual movie configuration is shown in Table \ref{tab:pc}. 

\begin{tcolorbox}
\textbf{Observation 3.} {\ours} is rational and interpretive on planning the blockbuster movie.
\end{tcolorbox}

In the planned movie configuration, we get a \$654.52 million on gross which is higher than the gross in original movie, \$623.27 million. All the members in the original movie are selected by {\ours}. Besides the actual members, we plan one more actor, actress and writer and two more directors which are shown with underscores.
For actors, we selected Sebastian Stan. We find that the reason why Sebastian Stan was selected is that he has a strong collaboration with Chris Evans as they collaborated twice and he participated in Marvel's ``Captain America'' series of movies, which share the same genre with ``The Avengers''. 
For actress, Elizabeth Olsen was selected. She has a strong collaboration with others, like Scarlett Johansson, Cobie Smulders and Chris Evans, because they participated in the movie ``Captain America: The Winter Soldier''. Additionally, she has participated in many the ``Action'', ``Adventure'' and ``Sci-Fi'' type movies, as she participated in the same genre movie ``Godzilla''. 
For writer, Joe Simon was selected. He is the writer in ``Captain America'' series of movies. And he collaborated with Chris Evans twice as well. 
For director, Jon Favreau was planned, as he is the director for the ``Iron Man'' series of movie. He collaborated with Robert Downey Jr. three times. Michael Bay, who is the director for the ``Transformers'' movie series, has collaborated with Scarlett Johansson in the movie ``The Island''. Besides, ``Iron Man'' series and ``Transformers'' series have the same genre as the ``The Avengers'' series. 
This planned configuration shows the effectiveness of our model, because it has a higher gross than the actual configuration and even if some casts was mistaken by us, all the planned casts are reasonable. 

\begin{table}[hbt]
\centering
\begin{tabular}{ |c|ccc|}
\hline
 \multicolumn{4}{|c|}{\textbf{Planned movie Configuration}} \\
  \hline
  \multirow{1}{*}{\textbf{Gross}} & \multicolumn{3}{c|}{ $\$654.52$}\\
  \hline
  \multirow{3}{*}{\textbf{Actor}} & \multicolumn{3}{c|}{ Robert Downey Jr., \underline{Sebastian Stan}, Tom Hiddleston}\\& \multicolumn{3}{c|}{Chris Hemsworth, Clark Gregg, Mark Ruffalo}\\ & \multicolumn{3}{c|}{Chris Evans, Jeremy Renner}\\
  \hline
  \multirow{2}{*}{\textbf{Actress}} & \multicolumn{3}{c|}{ \underline{Elizabeth Olsen}, Scarlett Johansson, Cobie Smulders}\\& \multicolumn{3}{c|}{Gwyneth Paltrow, Tina Benko, M'laah Kaur Singh}\\
  \hline
    \multirow{1}{*}{\textbf{Writer}} & \multicolumn{3}{c|}{ Joss Whedon, \underline{Joe Simon}, Zak Penn}\\
  \hline
    \multirow{1}{*}{\textbf{Director}} & \multicolumn{3}{c|}{ \underline{Jon Favreau}, Joss Whedon, \underline{Michael Bay}}\\
    \hline 
    \multicolumn{4}{c}{  } \\
      \hline 
\multicolumn{4}{|c|}{\textbf{Actual movie Configuration}} \\
  \hline
  \multirow{1}{*}{\textbf{ Gross}} & \multicolumn{3}{c|}{ $\$623.27$}\\
  \hline
    \multirow{3}{*}{\textbf{Actor}} & \multicolumn{3}{c|}{ Robert Downey Jr., Tom Hiddleston}\\& \multicolumn{3}{c|}{ Chris Hemsworth, Clark Gregg, Mark Ruffalo}\\& \multicolumn{3}{c|}{ Chris Evans, Jeremy Renner}\\
  \hline
 \multirow{2}{*}{\textbf{Actress}} & \multicolumn{3}{c|}{Scarlett Johansson, Cobie Smulders}\\& \multicolumn{3}{c|}{Gwyneth Paltrow, Tina Benko, M'laah Kaur Singh}\\
  \hline
    \multirow{1}{*}{\textbf{Writer}} & \multicolumn{3}{c|}{ Joss Whedon, Zak Penn}\\
  \hline
    \multirow{1}{*}{\textbf{Director}} & \multicolumn{3}{c|}{ Joss Whedon}\\
  \hline     
\end{tabular}
\caption{The planned and actual movie configuration of movie ``The Avengers''}
\label{tab:pc}
\end{table}

\section{Related Work} \label{sec:relate}
We have clearly illustrated the significant differences of the BP problem from the existing works in Section \ref{sec:introduction}. In this section, we provide a brief review of recent developments on related works. 

\noindent\textbf{Movie Gross Prediction} People use different resource of information to predict the movie gross. Mesty\'an and Yasseri \cite{mestyan2013early} used the knowledge base, Wikipedia, to predict the movie box office. Joshi et al. \cite{joshi2010mov} use the sentiment analysis on movie reviews to predict the movie gross. The recent analysis of the movie gross was done through social media, like Twitter and YouTube \cite{apala2013predict}\cite{doshi2010predicting}. 

\noindent\textbf{Viral Marketing} This problem focuses on finding a small set of seed nodes in a social network that maximizes the spread of influence. Kempe et al. \cite{kempe2003maximizing}, \cite{kempe2005influential} first proposed two basic diffusion models, namely independent cascade model(IC) and linear threshold model(LT). These two models set the foundation of almost all existing algorithms finding seed in social networks \cite{chen2009efficient}. The major drawback of their algorithm is that its inefficiency and ineffectiveness to the large networks. Later, Chen \cite{chen2009approximability} proposed a greedy algorithm to approximate the influence regions of nodes. However, when the scales beyond million-sized graphs, greedy algorithm becomes unfeasible. Chen et al. proposed to use local directed acyclic graphs to explore a large-scale influence maximization algorithm \cite{chen2010scalable}. 

\noindent\textbf{Team Formation} Lappas et al. first proposed this problem \cite{lappas2009finding}. They described an approach that defined the total communication cost among the social relationships to select a subset of experts to form a qualified team for certain projects. Recently, Nikolaev et al. proposed the EngTFP problem to find the subset of users that was the most important for keeping the whole user base together \cite{nikolaev2016engagement}. Different from those two works that find a subset of users to form a qualified team for certain projects, several recent works focus on training the team members \cite{zhang2017enterprise} \cite{bahargam2017team}. Their motivation is to build a team so that teammates can benefit from interaction to improve their skills.

\section{Conclusion}
In this work, we studied the Blockbuster Planning (BP) problem where professional movie planning are made by exploring the accumulated knowledge in the online movie knowledge library. A novel movie planning framework named {\ours} is introduced, where we first build the gross estimation function by analyzing and investigating the real-world online movie library dataset. The weights of the movie factors learned by the gross estimation are easily interpretable, and can be directly applied to the objective function for blockbuster planning. The {\ours} framework is optimized to maximize the movie gross as well as the {\cast} preference simultaneously. 
In addition, the limited budget is used as a hard constraint for the objective function to guarantee the plan achievement. Extensive experiments have been done on the real-world dataset to demonstrate the effective and advantages of the proposed framework in addressing the {\problem} problem. 

\section{Acknowledgements}
This work is supported in part by NSF through grants IIS-1526499, IIS-1763325, and CNS-1626432, and NSFC 61672313.
This work is also partially supported by NSF through grant IIS-1763365 and by FSU through the startup package and FYAP award.



\bibliographystyle{IEEEtran} 
\bibliography{00-mainsigkddExp}

\end{document}